\documentclass[10pt,twocolumn,letterpaper,dvipsnames]{article}

\usepackage[dvipsnames]{xcolor}
\usepackage{format}
\usepackage{times}
\usepackage{epsfig}
\usepackage{graphicx}
\usepackage{amsmath}
\usepackage{amssymb}
\usepackage{subcaption}
\usepackage{float}
\usepackage{dblfloatfix}
\usepackage{multirow}
\usepackage{color,soul}
\usepackage[accsupp]{axessibility}

\usepackage[pagebackref=true,breaklinks=true,letterpaper=true,colorlinks,bookmarks=false]{hyperref}

\finalcopy % *** Uncomment this line for the final submission

\begin{document}
\newcommand{\image}{\mathbf{x}}
\newcommand{\latent}{\mathbf{z}}
\newcommand{\camera}{\mathbf{c}}
\newcommand{\params}{\theta}
\newcommand{\jason}[1]{\textcolor{blue}{Jason: #1}}
\newcommand{\fereshteh}[1]{\textcolor{violet}{Fereshteh: #1}}
\newcommand{\kosta}[1]
{\textcolor{red}{\bf Kosta: #1}}
\newcommand{\marcus}[1]{\textcolor{orange}{Marcus: #1}}
\newcommand{\score}{\mathbf{s}_\params}
\newcommand{\webpage}{\url{https://yorkucvil.github.io/Photoconsistent-NVS/}}

\newcommand{\dir}{\mathbf{d}}
\newcommand{\ray}{\mathbf{r}}
\newcommand{\rays}{\mathcal{R}}
\newcommand{\point}{\mathbf{p}}
\newcommand{\fundamental}{\mathbf{F}}
\title{Long-Term Photometric Consistent Novel View Synthesis with Diffusion Models}

\author{Jason J. Yu$^{1,2}$, Fereshteh Forghani$^{1}$, Konstantinos G. Derpanis$^{1,2}$, Marcus A. Brubaker$^{1,2}$\\
$^{1}$York University, $^{2}$Vector Institute for AI\\
{\tt\small \{jjyu,forghani,kosta,marcus.brubaker\}@yorku.ca} \\
\webpage
}

%%%%%%%%% Main document
\twocolumn[{%
\renewcommand\twocolumn[1][]{#1}%
\maketitle
    \begin{center}
        \centering
        \captionsetup{type=figure}
        \includegraphics[width=1.0\textwidth]{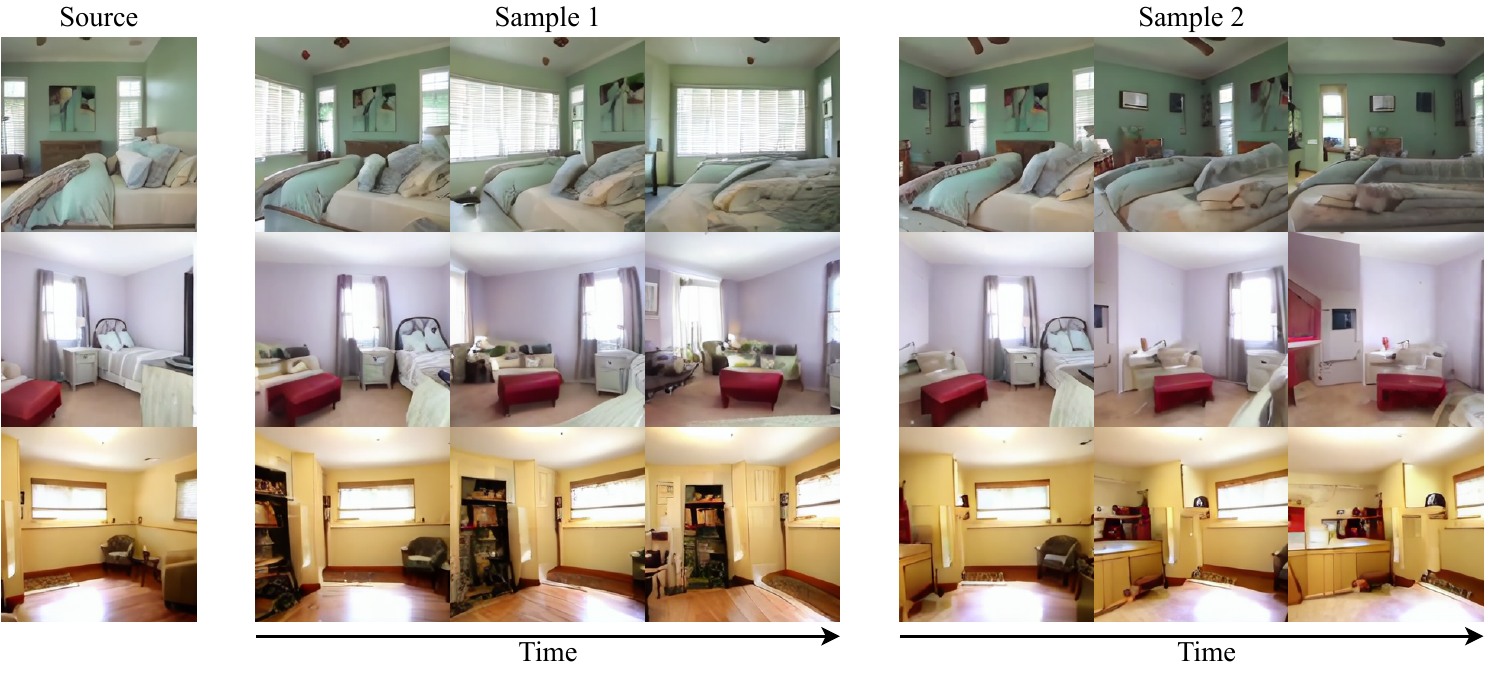}\vspace{-0.3cm}
        \captionof{figure}{
        Given a single source view, our model allows us to sample multiple plausible sets of views over a camera trajectory.
        Here, we show two samples (middle and right) of a sequence using the three source views (left).
        Our method is able to maintain consistency between observed regions, while plausibly extrapolating unseen regions.
        Notice that the final frames reveal regions that are largely unseen in the source view, and show different plausible appearances in each sample.
        }
        \label{fig:teaser}
    \end{center}%
}]

\begin{abstract}
Novel view synthesis from a single input image is a challenging task, where the goal is to generate a new view of a scene from a desired camera pose that may be separated by a large motion.
The highly uncertain nature of this synthesis task due to unobserved elements within the scene (i.e. occlusion) and outside the field-of-view makes the use of generative models appealing to capture the variety of possible outputs.
In this paper, we propose a novel generative model capable of producing a sequence of photorealistic images consistent with a specified camera trajectory, and a single starting image.
Our approach is centred on an autoregressive conditional diffusion-based model capable of interpolating visible scene elements, and extrapolating unobserved regions in a view, in a geometrically consistent manner.
Conditioning is limited to an image capturing a single camera view and the (relative) pose of the new camera view.
To measure the consistency over a sequence of generated views, we introduce a new metric, the \textit{thresholded symmetric epipolar distance} (TSED), to measure the number of consistent frame pairs in a sequence.
While previous methods have been shown to produce high quality images and consistent semantics across pairs of views, we show empirically with our metric that they are often inconsistent with the desired camera poses.
In contrast, we demonstrate that our method produces both photorealistic and view-consistent imagery.
\end{abstract}
\vspace{-0.7cm}
\section{Introduction}
Novel view synthesis (NVS) methods are generally tasked with generating new scene views, given a set of existing views. NVS has a long history in computer vision \cite{chen1993view,laveau19943,avidan1997novel} and has recently seen a resurgence of interest with the advent of NeRFs \cite{mildenhall2021nerf,yu2021pixelnerf,verbin2022ref}.
Most current approaches to NVS (\eg NeRFs) focus on problem settings where generated views remain close to the input and whose content is largely visible from some subset of the given views.
This restricted setting makes these methods amenable to direct supervision.
In contrast, we consider a more extreme case, where a single view is given as input, and the goal is to generate plausible image sequence continuations from a trajectory of provided camera views.
By plausible, we mean that visible portions of the scene should evolve in a 3D consistent fashion, while previously unseen elements (\ie regions occluded or outside of the camera field-of-view) should appear harmonious with the scene.
Moreover, regions not visible in the input view are generally highly uncertain; so, there are a variety of plausible continuations that are valid.

To address this challenge, we propose a novel NVS method based on denoising diffusion models \cite{ho2020denoising} to sample multiple, consistent novel views.
We condition the diffusion model on both the given view, and a geometrically informed representation of the relative camera settings of both the given and target views.
The resulting model is able to produce multiple plausible novel views by simply generating new samples from the model.
Further, while the model is trained to generate a single novel view conditioned on an existing view and a target camera pose, we demonstrate that this model can generate a sequence of plausible views, including final views with little or no overlap with the starting view.
Fig.\ \ref{fig:teaser} shows the outputs of our model for several different starting views, with two samples of plausible sets of views.

Existing NVS techniques have been evaluated primarily in terms of generated image quality (\eg with Fréchet Inception Distance (FID) \cite{NIPS2017_8a1d6947}) but have generally ignored measuring consistency with the camera poses.
Based on the epipolar geometry defined by relative camera poses \cite{hartley2003multiple}, we introduce a new metric which directly evaluates the geometric consistency of generated views independently from the quality of generated imagery.
The proposed metric does not require any knowledge of scene geometry, making it widely applicable even on purely generated images.
We evaluate the proposed method on both real and synthetic datasets in terms of both generated image quality and geometric consistency.
Further, previous work only evaluates performance based on in-distribution camera trajectories.
Here, we evaluate the generalization ability of extant models and our own by generating sequences based on novel trajectories (\ie trajectories that differ significantly from those in the training data).
\section{Related Work}
NVS has been long studied in computer vision (\eg \cite{chen1993view,laveau19943,avidan1997novel}), and a full review is out of scope for this paper.
NVS methods can largely be categorized as those which focus on \emph{view interpolation}, where generated views remain close to the given views, and \emph{view extrapolation}, where the generated field-of-view may contain large amounts of novel content.
Many current view interpolation methods are based on NeRFs \cite{mildenhall2021nerf,yu2021pixelnerf}, which leverage neural-network representations of radiance fields fit to the observed images.
Others attempt to directly regress novel views \cite{sajjadi2022scene} from a set-encoded representation of the given views.
Alternatively, if depth information is available, images can be back-projected into 3D, and missing regions inpainted \cite{koh2022simple}.
We focus on view extrapolation NVS where significant portions of the generated images are not visible in the inputs.

View extrapolation methods are largely built on probabilistic approaches to capture the high degree of uncertainty.
GAUDI \cite{bautista2022gaudi} learns a latent variable model of entire 3D scenes represented as a neural radiance field and then estimates the latents given observed images.
However, the estimated scene representation often has a limited spatial extent, which is in contrast to image-to-image methods \cite{liu2021infinite} which may extend indefinitely.
GeoGPT \cite{rombach2021geometry} uses an autoregressive likelihood model to sample novel views conditioned on a single source view.
In contrast, we use a latent diffusion model \cite{NEURIPS2021_701d8045, rombach2022high}, and investigate sequential view generation.
LookOut \cite{ren2022look} extends GeoGPT \cite{rombach2021geometry} to generate sequences of views along a trajectory while conditioning on up to two previous views.
To enforce consistency, LookOut requires a post-processing step that uses generated outputs as additional conditioning.
In contrast, our model is conditioned on a single view, and does not require additional post-processing to achieve consistency.
A closely related method \cite{watson2022novel} also formulates a diffusion model for NVS; however, it was only applied to simple scenes (\ie isolated objects) with constrained camera poses.
Here, we consider view extrapolation on real indoor scenes with complex geometry, and without constraints on camera motion.

Conditional generative models are a common approach for view synthesis \cite{rombach2021geometry,ren2022look}, image editing \cite{meng2022sdedit}, and video prediction \cite{lee2018savp}. 
Recent years has seen significant progress in generative modelling \cite{kingma2013auto,salimans2017pixelcnn++,ho2020denoising,goodfellow2020generative,esser2021taming} with diffusion models \cite{rombach2022high,song2020score,ho2020denoising} showing promise in many tasks, \eg text-to-image generation \cite{rombach2022high,saharia2022photorealistic} and video modeling \cite{ho2022video}.
In our problem, we utilize latent diffusion models \cite{NEURIPS2021_701d8045, rombach2022high}, which first compress high dimensional images with an autoencoder and discourage the diffusion model from expending capacity on modeling imperceptible details.
The resulting model is more efficient, and uses less computation during training and inference.

Generative methods and 3D capable models are currently a very active research topic and there have been other highly related concurrent works investigating pose-conditional diffusion models.
RenderDiffusion \cite{anciukevivcius2023renderdiffusion} uses an explicit 3D tri-plane representation \cite{chan2022efficient} for object-centric NVS, and relies on score-distillation \cite{poole2022dreamfusion} for 3D regularization, rather than relying on multi-view training data.
In contrast to our method, RenderDiffusion focuses on object-centric NVS, and utilizes a 3D representation with limited spatial extent, while we focus on extrapolating scenes.
The pose-guided diffusion model from Tseng et al. \cite{tseng2023consistent} is very similar to our method but uses a cascade diffusion model \cite{ho2021cascaded}, and only investigates performance on in-distribution trajectories.
In contrast, our method uses a latent diffusion model, and investigates generalization to out-of-distribution trajectories.

\section{Technical Approach}

\subsection{Background: Diffusion Models}
% A full review of diffusion models is beyond the scope of this work.
Here, we provide a brief introduction of diffusion models to ground the following developments but refer interested readers to a recent detailed review \cite{DIFFUSIONREVIEWARTICLE}.
Diffusion models are a class of generative models where sampling is performed by reversing a stochastic diffusion process \cite{ho2020denoising,song2020score}.
The forward process is fixed, typically Gaussian, and discretized into $t \in 1,...,T$ timesteps which are defined recursively as
\begin{align}
    q(\image_{t} | \image_{t-1}) &= \mathcal{N}(\image_t;\sqrt{1-\beta_t}\image_{t-1},\beta_t\mathbf{I}), \label{eq:forward_noise}
\end{align}
where $\image_0$ is a sample from the data distribution of interest, $q(\image_0)$, $\mathbf{I}$ is an identity matrix, and the values of $\beta_t$ are dependent on the particular forward process used.
Repeatedly applying Eq.\ \ref{eq:forward_noise} adds Gaussian noise with $\image_T$ approximately normally distributed for large values of $T$.
The reverse process is parameterized by $\theta$ and takes the form of a Gaussian:
\begin{align}
    p_\theta(\image_{t-1}|\image_{t}) &= \mathcal{N}(\image_{t-1};\mu_\theta(\image_{t},t),\Sigma_\theta(\image_{t},t)), \label{eq:reverse_process}
\end{align}
where the variance, $\Sigma_\theta(\image_{t-1},t)$, is generally set as constant.
Here, $\image_{t-1}$ is expressed using $\epsilon_\theta(\image_t,t)$ which is implemented as a neural network:
\begin{align}
    \image_{t-1} = \frac{1}{\sqrt{\alpha_t}}\left(\image_t - \frac{1-\alpha_t}{\sqrt{1-\Bar{\alpha}_t}}\epsilon_\theta(\image_t,t)\right) + \sigma_t\mathbf{z}_t,
\end{align}
where $\alpha_t = 1-\beta_t$, $\Bar{\alpha}_t = \prod_{s=1}^t \alpha_s$, and $\mathbf{z}_t \sim \mathcal{N}(0,I)$.
The function $\epsilon_\theta(\image_t,t)$ is referred to as the score function and can be interpreted as a noise estimator which can be used to denoise $\image_t$ to produce $\image_{t-1}$.
Training is performed using denoising score matching \cite{vincent2011connection}:
\begin{align}
    \mathcal{L} = \mathbb{E}_{\image_0,t,\epsilon} \left[ || \epsilon - \epsilon_\theta(\sqrt{\Bar{\alpha}_t}\image_0 + \sqrt{1-\Bar{\alpha}_t}\epsilon,t) ||^2 \right].
\end{align}

Samples can be drawn from the model by initializing $\image_T$ with Gaussian noise, and iteratively applying the learned reverse process given in Eq.\ \ref{eq:reverse_process}.
The model is made conditional by providing additional inputs to the score function, $\epsilon_\theta(\image_t,t)$.
Due to the redundant and high dimensional nature of images, it is beneficial to first reduce their dimensionality.
There are several ways to approach the dimensionality reduction task \cite{guth2022wavelet,ho2021cascaded,NEURIPS2021_701d8045,rombach2022high}.
Here, we use a latent diffusion model \cite{NEURIPS2021_701d8045,rombach2022high} that first transforms an image, $\image$, into a latent representation, $\latent$, with a learned autoencoder, $\latent = \text{E}(\image)$.
The diffusion model is then learned in the latent space, $\latent$, and images are recovered by using the corresponding decoder, $\image = \text{D}(\latent)$.
Critically for us, the learned latent representation can maintain the spatial structure of the image, \eg through the use of a convolutional encoder architecture.

\begin{figure}
    \centering
    \includegraphics[width=\linewidth]{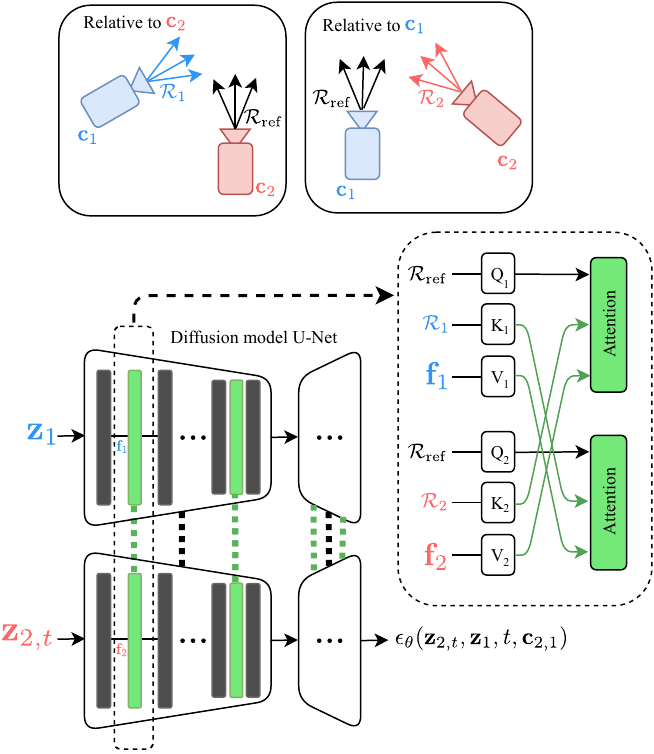}\vspace{-0.35cm}
    \caption{
    An overview of our model with two streams coupled with cross-attention.
    Our diffusion model is implemented as a two-stream U-Net \cite{ronneberger2015u}, where latent representations for the given view, ${\color{blue}\latent_1}$ (\textcolor{blue}{blue}) and the generated view at diffusion step $t$, ${\color{red}\latent_{2,t}}$ (\textcolor{red}{red}), are processed by separate streams consisting of spatial layers with shared parameters (black).
    The latent of the given view, ${\color{blue}\latent_1}$, is used to condition the score of ${\color{red}\latent_{2,t}}$, and the camera poses are $\camera_1$ and $\camera_2$.
    Both streams are conditioned on the noise variance, which is omitted for clarity. The two streams communicate via cross-attention layers (\textcolor{Green}{green}). The queries are augmented with rays in a canonical reference frame, $\mathcal{R}_\text{ref}$.
    The keys, $\text{K}_1$ and $\text{K}_2$, are augmented with ray information, ${\color{blue}\mathcal{R}_\text{1}}$ and ${\color{red}\mathcal{R}_\text{2}}$, respectively, which are each localized in the reference frame of the opposite view, ${\color{red}\mathbf{c}_\text{2}}$ and ${\color{blue}\mathbf{c}_\text{1}}$, illustrated on the top.
    The inset on the middle-right illustrates the cross-attention layer, where ${\color{blue}\mathbf{f}_1}$ and ${\color{red}\mathbf{f}_2}$ are incoming features.
    }
    \label{fig:cross_attn}
    \vspace{-1em}
\end{figure}
\subsection{Novel View Synthesis with Diffusion Models}
\label{sec:method}
We now describe how we use a diffusion model to sample multiple plausible views in novel view synthesis.
A conditioning image, $\image_1$, is first mapped into the latent space, $\latent_1 = \text{E}(\image_1)$, and then is used to condition the distribution over the latent representation of the desired view:
\begin{align}
    p_\params(\latent_2 | \latent_1,\camera_{2,1}), \label{eq:cond_prob}
\end{align}
where $\camera_{2,1}$ is the relative camera pose between the source and target views.
The distribution is estimated using a diffusion model \cite{song2020score} with score function $\epsilon_\theta(\latent_{2,t},\latent_1,t,\camera_{2,1})$, where $\latent_{2,t}$ is the value of $\latent_{2}$ at diffusion step $t$.
The novel view, $\image_2$, is then decoded from the sampled latent representation: $\image_2 = \text{D}(\latent_{2,T})$.

To obtain views along a trajectory with our model we generate them in sequence.
Ideally, these would be sampled using the distribution conditioned on all previously generated views:
\begin{align}
    \latent_{i+1} \sim p(\latent_{i+1}|\latent_{0},\dotsc,\latent_{i},\camera_{i+1,0},\dotsc,\camera_{i+1,i})\ . \label{eq:dist-full}
\end{align}
We approximate this by assuming a Markov relationship between views in the sequence.
That is, given an initial image, $\image_0$, samples in the sequence of length $L$ are obtained by encoding the initial image, $\latent_0 = \text{E}(\image_0)$, and recursively sampling from:
\begin{align}
    \latent_{i+1} \sim p(\latent_{i+1}|\latent_{i},\camera_{i+1,i}),
\end{align}
with the final image decoded from the sampled latent representation: $\image_{L-1} = \text{D}(\latent_{L-1})$.
We structure our model specifically for NVS, by equipping it with a specialized representation for relative camera geometry, and a two-stream architecture.

Reasoning about novel views requires knowledge of geometric camera information.
To provide this information we augment the input of the score function with a representation of the camera rays for the conditioning and generated views \cite{yifan2022input,sajjadi2022scene}.
Our camera model is defined by the intrinsic matrix, $\mathbf{K}$, and the extrinsics, $\camera = \left[\mathbf{R}|\mathbf{t}\right]$, where $\mathbf{R}$ and $\mathbf{t}$ are the 3D rotation and translation components, respectively.
Given the projection matrix of a camera, $\mathbf{P} = \mathbf{K}[\mathbf{R}|\mathbf{t}]$, the camera center is computed as $\boldsymbol{\tau} = -\mathbf{R}^{-1}\mathbf{t}$.
The direction of the camera ray at pixel coordinates $(u,v)$ is given by:
\begin{align}
    \Bar{\dir}_{u,v} = \mathbf{R}^{-1} \mathbf{K}^{-1}
    \begin{bmatrix}
        u & v & 1
    \end{bmatrix}^\top,
\end{align}
which is then normalized to unit length to obtain $\dir_{u,v}$.
Finally, before being used as conditioning for the diffusion model, the ray direction is concatenated with the camera center, $\ray_{u,v} = [\dir_{u,v},\boldsymbol{\tau}]$, and frequency encoded \cite{vaswani2017attention} :
\begin{align}
    \mathcal{R} = \left[ \sin(f_1 \pi \ray), \cos(f_1 \pi \ray), \dotsc, \sin(f_K \pi \ray), \cos(f_K \pi \ray) \right],
\end{align}
where $K$ is the number of frequencies, $f_k$ are the frequencies which increase proportionally to $2^k$, and the sinusoidal functions are applied element-wise.

The standard architecture for a score function is a U-Net architecture \cite{ronneberger2015u}.
Here, we base our architecture on the Noise Conditional Score Network++ (NCSN++) architecture \cite{song2020score}, with a variance exploding forward process.
We modify this backbone architecture to incorporate the ray representation and the conditioned view.
Inspired by video diffusion models \cite{ho2022video}, we propose a two-stream architecture using two U-Nets with shared weights to process the novel view, $\image_{2,t}$, and conditioning view, $\image_{1}$.
These networks communicate with one another exclusively via cross-attention layers, which are inserted after every spatial attention layer.
We also augment the queries and keys of the attention with camera pose information.
The output of the novel view stream is used as the output of the score function, $\epsilon_\theta(\latent_{2,t},\latent_1,t,\camera_{2,1})$.
In short, the model contains a stream for each view, and couples them using augmented cross-attention.
Our architecture is illustrated in Fig.\ \ref{fig:cross_attn} and more details are given in Appendix\ \ref{sec:arch}.

\subsection{Thresholded Symmetric Epipolar Distance (TSED)}
\label{sec:tsed}
Existing evaluation metrics for NVS primarily focus on the view interpolation case and are based on notions of reconstruction (\eg PSNR and LPIPS) or general image quality (\eg FID); however, reconstruction metrics are inapplicable to view extrapolation, where there is no reasonable expectation of a single ground truth output.
General image quality metrics are relevant for view extrapolation but existing measures like FID are insensitive to the accuracy of the geometry.
That is, generated images can completely ignore the required camera pose and still achieve excellent FID.
To address this issue recent work \cite{watson2022novel} proposed a metric that is sensitive to accurate camera geometry, but the evaluation involves fitting a NeRF \cite{mildenhall2021nerf} to multiple generated images, and measuring consistency as the FID of unseen interpolated views; however, this evaluation is complex, excessively expensive to compute, and difficult to interpret.
Here, we propose the Thresholded Symmetric Epipolar Distance (TSED) as a new lightweight metric for measuring geometric consistency of NVS models.

Our metric is motivated by two consistency criteria.
First, the appearance of objects should remain stable between views, and should contain image features that can be identified and matched.
Second, these matched features should respect epipolar constraints \cite{hartley2003multiple}, given by the desired relative camera pose.
With the camera poses used to condition the generation of the novel view, we compute the fundamental matrix, $\fundamental$, which, given a feature point $\point$ in one image, allows us to define the epipolar line $\point'^\top \fundamental \point=0$ on which its corresponding feature $\point'$ should lie.
We define the symmetric epipolar distance (SED) of corresponding points $\point$ and $\point'$ as:
\begin{align}
    \text{SED}(\point,\point',\fundamental) = \frac{1}{2}\left[d(\point',\fundamental\point) + d(\point,\fundamental^\top \point')\right],
\end{align}
where $d(\point',\fundamental\point)$ is the minimum Euclidean distance between point $\point'$ and the epipolar line induced by $\fundamental\point$.
(We note this definition of SED is similar in spirit but slightly different than those found in some standard references.)
Given a set of feature correspondences, $M = \{(\point_1,\point_1'),\dotsc,(\point_n,\point_n')\}$, between two views (\eg computed with SIFT \cite{lowe1999object}) we define the pair of images to be consistent if there are a sufficient number of matching features, \ie $n \geq T_\text{matches}$, and the median SED over $M$ is less than $T_\text{error}$.
The median is chosen to mitigate the influence of incorrect correspondences.
The threshold $T_\text{matches}$ makes the metric robust against image pairs with few matches as this likely indicates a low-quality generation, assuming the scenes are not largely textureless and the cameras do not undergo an extreme viewpoint change.
The use of epipolar geometry here is key as it does not require knowledge of the scene geometry or scale.
It should be noted that using epipolar geometry results in TSED having lower sensitivity to errors when most of the epipolar lines have a similar orientation, because SED for a match is insensitive to errors in 2D correspondence that parallel to the epipolar line.
An empirical sensitivity analysis of TSED is provided in Appendix\ \ref{sec:sensitivity}.
Given a NVS model, we evaluate it by generating sequences of images and computing which fraction of neighbouring views are consistent.
We use $T_\text{matches}=10$ and explore consistency as a function of different values of $T_\text{error}$ in our experiments.

\section{Experiments}
We evaluate and compare to extant methods with a focus on \emph{both} independent image quality and consistency across views.
We conduct an ablation study on CLEVR \cite{johnson2017clevr}, a synthetic dataset, to validate the various components of our model (Sec.\ \ref{sec:ablations}).
We further demonstrate the capabilities of our model using RealEstate10K \cite{realestate46965}, a large dataset of real indoor scenes, Matterport3D \cite{chang2017matterport3d}, a small dataset of building-scale textured meshes, and compare our method with two strong baselines (Sec.\ \ref{sec:gt-experiments}): GeoGPT \cite{rombach2021geometry} and LookOut \cite{ren2022look}.

\subsection{Experimental Setup}\label{sec:exp_setup}
For our experiments, we implement our model using a latent diffusion model (LDM) \cite{rombach2022high} with a VQ-GAN \cite{esser2021taming} as the latent space autoencoder, and a modified architecture as described in the previous section.
During inference, we sample with ancestral sampling using a predictor-corrector sampler \cite{song2020score}.
Training requires pairs of images along with camera intrinsics, and relative extrinsics.
For evaluation we use the CLEVR \cite{johnson2017clevr}, RealEstate10K \cite{realestate46965}, and Matterport3D (MP3D) \cite{chang2017matterport3d} datasets.

CLEVR \cite{johnson2017clevr} is a synthetic dataset consisting of scenes of simple geometric primitives with various materials placed on top of a matte grey surface.
We repurpose the Blender based pipeline to uniformly scatter the primitives in the center of the scene in an $8\times8$ Blender unit area, and render views from a slightly elevated position to prevent the camera from being placed inside an object.
The initial camera position is chosen uniformly in the same area that the objects are placed, and oriented towards the center of the scene with a $[-20,20]$ degree jitter around the yaw axis.
For the second view, the camera is randomly translated $[-1,1]$ units along the ground plane, and jittered $[-20,20]$ degrees around the yaw axis.
Images are rendered at a resolution of $128\times128$.
The left most panel in Fig.\ \ref{fig:clevr-samples-traj} provides an example image.

RealEstate10K \cite{realestate46965} consists of publicly available real estate tour videos scraped from YouTube.
The videos are partitioned into disjoint sequences, and the camera parameters provided with the dataset were recovered using ORB-SLAM2 \cite{mur2015orb}.
The large amount of real, diverse, and structured environments available in RealEstate10K make it an ideal and commonly used dataset for NVS evaluation, including by the most relevant baselines \cite{rombach2021geometry,ren2022look}.
Following previous work \cite{ren2022look}, the videos are obtained at 360p, center cropped, and downsampled to $256\times256$.
One challenging aspect of using this dataset is the limited diversity in camera motions.
Many of the sequences consists of a simple forward motion that travels through and between rooms.
This gives us an opportunity to evaluate the generalization of the model to novel camera motions not present in the dataset.

Matterport3D \cite{chang2017matterport3d} consists of 90 indoor, building-scale environments that have been scanned using RGB-D sensors, and reconstructed as a textured mesh.
Following previous work \cite{ren2022look}, we convert the scenes into videos using an embodied agent in the Habitat \cite{savva2019habitat} simulation platform to navigate between two randomly chosen locations in the scene, and render each frame at a resolution of $256\times256$.
For each frame of the sequence, the agent chooses one of three actions: move forward, turn left, and turn right.
The limited actions that the agent can perform greatly reduces the diversity of camera motions, which is even more limited than those available in RealEstate10K.

For our evaluations, we compare our method with two recent state-of-the-art generative scene extrapolation methods.
GeoGPT \cite{rombach2021geometry} is an image-to-image NVS method, using a similar probabilistic formulation as our method.
Four variants were proposed with options to leverage monocular depth maps provided by MiDaS \cite{Ranftl2022}, and perform explicit warping of the source image.
For our evaluation, we use their model with implicit geometry and without access to depth maps as this is most similar to our proposed method, and can, similarly, be applied autoregressively to generate sequences. 
LookOut\footnote{An official public implementation is available without the pretrained weights on RealEstate10k. After email correspondence with the authors, we were unable to obtain the pretrained model. Reported results are based on a retrained model using the authors' publicly available code.}\ \cite{ren2022look} is an extension of GeoGPT with a focus on improving the generation of novel views over a long camera trajectory.
The model takes up to two input frames of a sequence, and uses the camera pose information to explicitly bias the attention layers inside the model.
The final LookOut model is fine-tuned with a form of simulated error accumulation \cite{liu2021infinite} to make the model robust to errors present in its inputs during autoregressive generation.
In our evaluation, we consider two variants of LookOut, one including this post-processing step (LookOut), and one without (LookOut-ne).
Note both GeoGPT and our model do not include a post-processing step and could potentially benefit from it.
For MP3D, we use the publicly available weights for LookOut.

\begin{table}
    \centering
    % \footnotesize
    \begin{tabular}{c|cc|c}
        \multirow{2}{*}{Method}  & \multicolumn{2}{c|}{Single view} & Last view\\
         & LPIPS $\downarrow$ & PSNR $\uparrow$ & FID $\downarrow$ \\
        \hline
        Naive concat  & 0.121 & 23.24 & 79.57 \\
        Two-Stream SC & - & - & 78.11 \\
        Two-Stream    & \textbf{0.112} & \textbf{24.20} & \textbf{76.85} \\
    \end{tabular}
    \vspace{-0.5em}
    \caption{Reconstruction metrics and FID for single view prediction and sequential prediction on CLEVR. \textit{Two-Stream} is our two stream model, \textit{Two-Stream SC} is our two-stream model sampled with stochastic conditioning, and \textit{Naive concat} is the naive variant where inputs are concatenated along the channel dimension. We evaluate the FID on the last generated image of a trajectory. Stochastic conditioning is only applicable with more than two generated views, no results are provided for this method on single view evaluations.}
    \label{tab:clevr-recon}
       \vspace{-10pt}
\end{table}
\begin{figure}
    \centering
    \includegraphics[width=\linewidth]{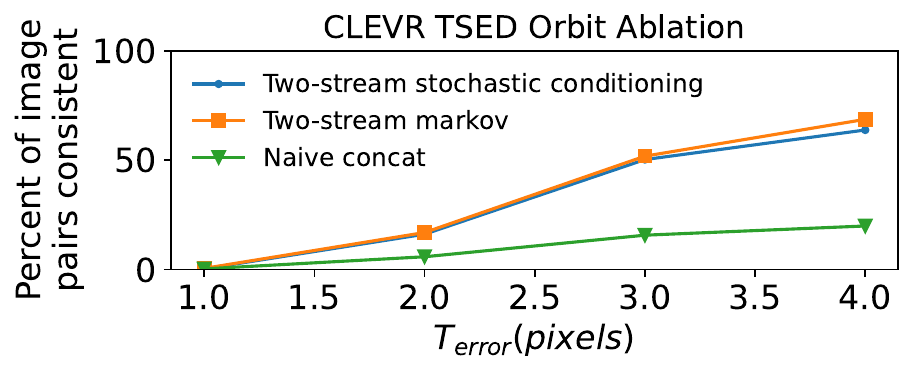}
    \vspace{-2em}
    \caption{Percent consistent image pairs computed with TSED on different variants of our model, and sampling, on CLEVR.
    }
    \vspace{-1em}
    \label{fig:clevr-tsed}
\end{figure}
In addition to our introduced consistency metric (Sec.\ \ref{sec:tsed}), we evaluate the quality of the generated images using standard image-centric metrics, specifically PSNR, LPIPS, and FID.
PSNR and LPIPS are standard full reference image reconstruction metrics used to evaluate differences between generated and ground truth views.
However, as the camera view changes significantly the space of plausible views increases dramatically and reconstruction metrics like PSNR and LPIPS become less relevant due to a lack of single ground truth reference.
While these metrics are not suitable for evaluating view extrapolation tasks \cite{teterwak2019boundless,rockwell2021pixelsynth}, they can still provide some sense of consistency for short-term generation, where uncertainty in the novel views is low.
FID \cite{NIPS2017_8a1d6947} is a standard reference-free metric for generative methods which measures sample quality of a set of i.i.d.\ samples, compared to a set of real samples.
While FID does not provide a measure of consistency between images, it gives a sense of the overall realism of the generated images.

\subsection{Ablations}
\label{sec:ablations}
\begin{figure}
    \centering
    \includegraphics[width=\linewidth]{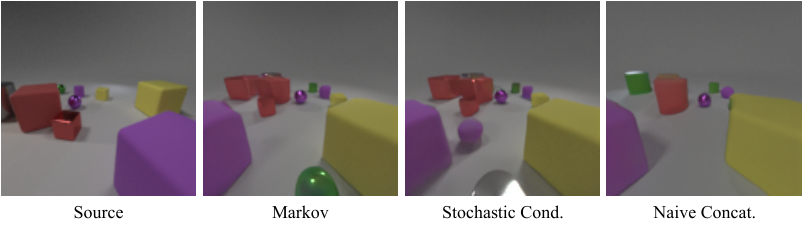}
    \vspace{-2em}
    \caption{Samples of the sixth generated frame from the initial image on the left. Note the small red cube visible in the initial image disappears for the naive model.}
    \label{fig:clevr-samples-traj}
\end{figure}
Here, we explore variations on model architecture and sampling, and compare performance.
First, we compare our two-stream architecture with a naive conditional diffusion model architecture, where both source and target views are concatenated to create a six channel image, and the model estimates the score for the target view.
The results are shown in Tab.\ \ref{tab:clevr-recon}, which shows clearly that our proposed architecture is effective.

We also explore an alternative strategy for sampling trajectories of novel views.
Previous work \cite{watson2022novel} proposed a heuristic for extending a single source view novel view diffusion model to use an arbitrary number of source views called \textit{stochastic conditioning}.
Given $m$ possible source views, each iteration of the diffusion sampling process is modified to be randomly conditioned on one of the $m$ views.
We consider this heuristic for generating sets of views, conditioning on up to two of the previous frames.
For these ablations, we sample ten images from a trajectory orbiting the center of the scene, using 100 different starting images.

We evaluate consistency using TSED; quantitative results are provided in Fig.\ \ref{fig:clevr-tsed}, and qualitative results are shown in Fig.\ \ref{fig:clevr-samples-traj}.
We find that the naive model can generate images where clearly visible objects may disappear, leading to less consistency qualitatively and quantitatively.
Sampling with stochastic conditioning is qualitatively similar to Markov sampling.
Quantitatively, stochastic conditioning is less consistent when $T_\text{error}$ is high, which is the result of fewer matches being made.
In general, recovering correspondences on CLEVR is challenging due to few distinct features.
Despite the challenges presented by this dataset, our metric is still able to provide a measure of consistency.
Overall, these results show that in contrast to previous work \cite{watson2022novel}, stochastic conditioning has no benefit to our approach and may actually hurt performance.
We also attempt to perform stochastic conditioning on RealEstate10K, but the images are qualitatively poor; results are available in the Appendix\ \ref{sec:sc}.

\subsection{Generation with In-Distribution Trajectories}
\label{sec:gt-experiments}
For our initial set of experiments on RealEstate10K and MP3D, we consider the generation of novel views along in-distribution trajectories.
To generate representative, in-distribution trajectories, given a start image, we randomly sample camera trajectories from the test set, as done in previous work \cite{ren2022look}.

\begin{table}
    \footnotesize
    \centering
    %\resizebox{\linewidth}{!}{
    \begin{tabular}{c|c|cc|cc}
        &\multirow{2}{*}{Method}  & \multicolumn{2}{c|}{Short-term} & \multicolumn{2}{c}{Long-term} \\
        & & LPIPS $\downarrow$ & PSNR $\uparrow$ & LPIPS $\downarrow$ & PSNR $\uparrow$ \\
        \hline
        \multirow{4}{*}{\rotatebox[origin=c]{90}{\resizebox{0.14\linewidth}{!}{RealEstate10K}}}&GeoGPT \cite{rombach2021geometry}  & 0.444 & 13.35 & 0.674 & 9.54 \\
        &LookOut-ne \cite{ren2022look} & 0.390 & 14.19 & 0.688 & 9.65 \\
        &LookOut \cite{ren2022look} & 0.378 & 14.43 & 0.658 & 10.51 \\
        &Ours    & \textbf{0.333} & \textbf{15.51} & \textbf{0.588} & \textbf{11.54} \\
        \hline\hline
        \multirow{2}{*}{\rotatebox[origin=c]{90}{\resizebox{0.06\linewidth}{!}{MP3D}}}&LookOut \cite{ren2022look} & 0.604 & 12.76 & 0.739 & 10.60 \\ 
        &Ours                       & \textbf{0.504} & \textbf{14.83} & \textbf{0.674} & \textbf{13.00}
    \end{tabular}
    %}
    \caption{
    RealEstate10K and MP3D reconstruction metrics with in-distribution trajectories.
    LookOut-ne refers to the LookOut method without the final error accumulation training step.
    }
       \vspace{-10pt}
    \label{tab:recon-gt-all}
\end{table}
\begin{table}
    \centering
    \footnotesize
    %\resizebox{\linewidth}{!}{
    \begin{tabular}{c|c|cc}
        & Method & Short-term FID $\downarrow$ & Long-term FID $\downarrow$ \\
        \hline
        \multirow{4}{*}{\rotatebox[origin=c]{90}{\resizebox{0.14\linewidth}{!}{RealEstate10K}}} &GeoGPT \cite{rombach2021geometry}  & \textbf{26.72} & \textbf{41.87} \\
        &LookOut-ne \cite{ren2022look} & 30.38 & 72.01 \\
        &LookOut \cite{ren2022look} & 28.86 & 58.12 \\
        &Ours    & 26.76 & 41.95 \\
        \hline\hline
        \multirow{2}{*}{\rotatebox[origin=c]{90}{\resizebox{0.06\linewidth}{!}{MP3D}}} &LookOut \cite{ren2022look}  & 80.97 & 132.36 \\
        &Ours                        & \textbf{73.16} & \textbf{100.99} \\
    \end{tabular}
    %}
    \caption{
    RealEstate10K and MP3D FIDs with in-distribution trajectories.
    FID scores between generated images at short-term and long-term generations, and a fixed set of randomly selected images from the test set.
    }
       \vspace{-10pt}
    \label{tab:FID-gt-all}
\end{table}
\begin{figure}
    \centering
    \begin{subfigure}[b]{0.49\linewidth}
        \centering
        \includegraphics[width=\linewidth]{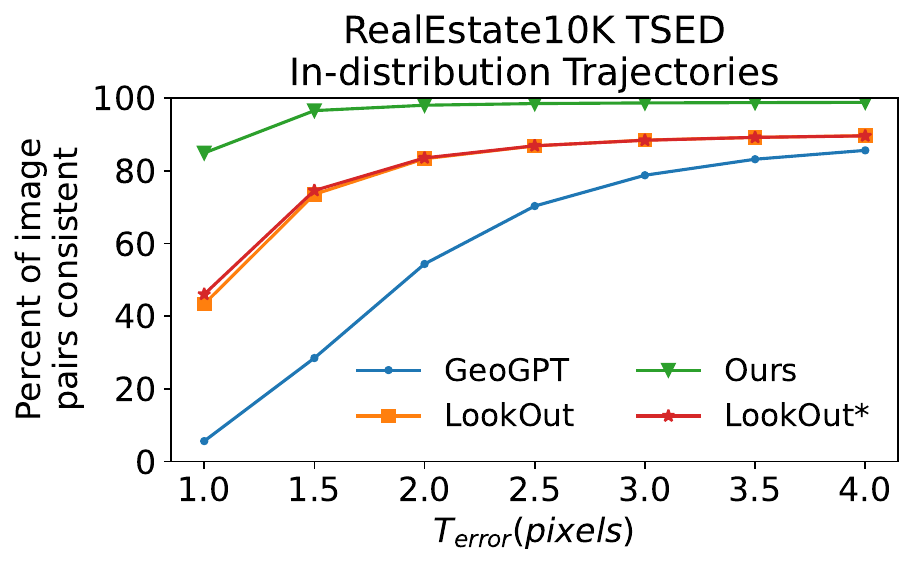}
        \vspace{-20pt}
        % \caption{Hop}
    \end{subfigure}
    \begin{subfigure}[b]{0.49\linewidth}
        \centering
        \includegraphics[width=\linewidth]{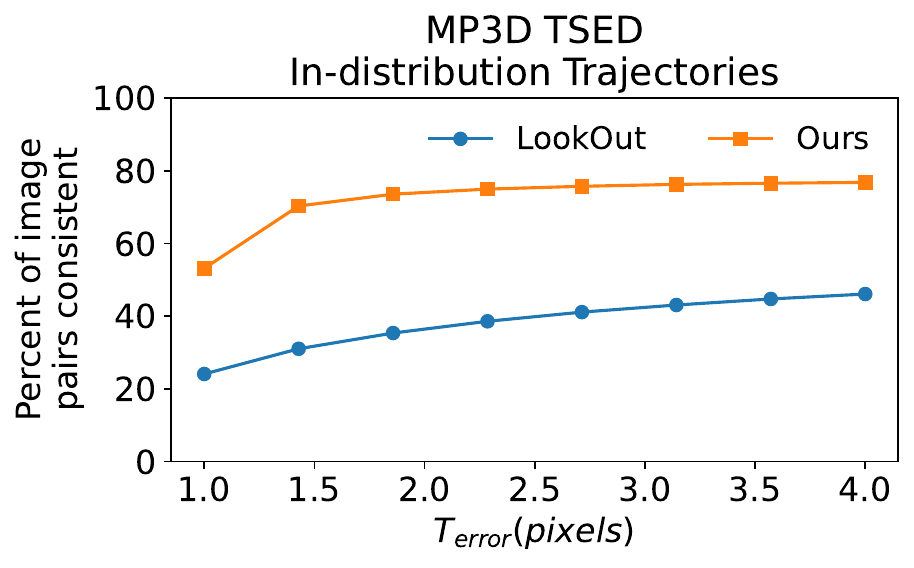}
        \vspace{-20pt}
        % \caption{Orbit}
    \end{subfigure}
    \caption{
    RealEstate10K TSED on in-distribution trajectories.
    Consistency is measured as the average percent of consistent image pairs in the generated sequences. We set $T_\text{matches} = 10$.
    }
    \label{fig:cons-gt-both}
    \vspace{-10pt}
\end{figure}
\begin{figure}
    \centering
    \includegraphics[trim={0 0 0 6em},width=\linewidth]{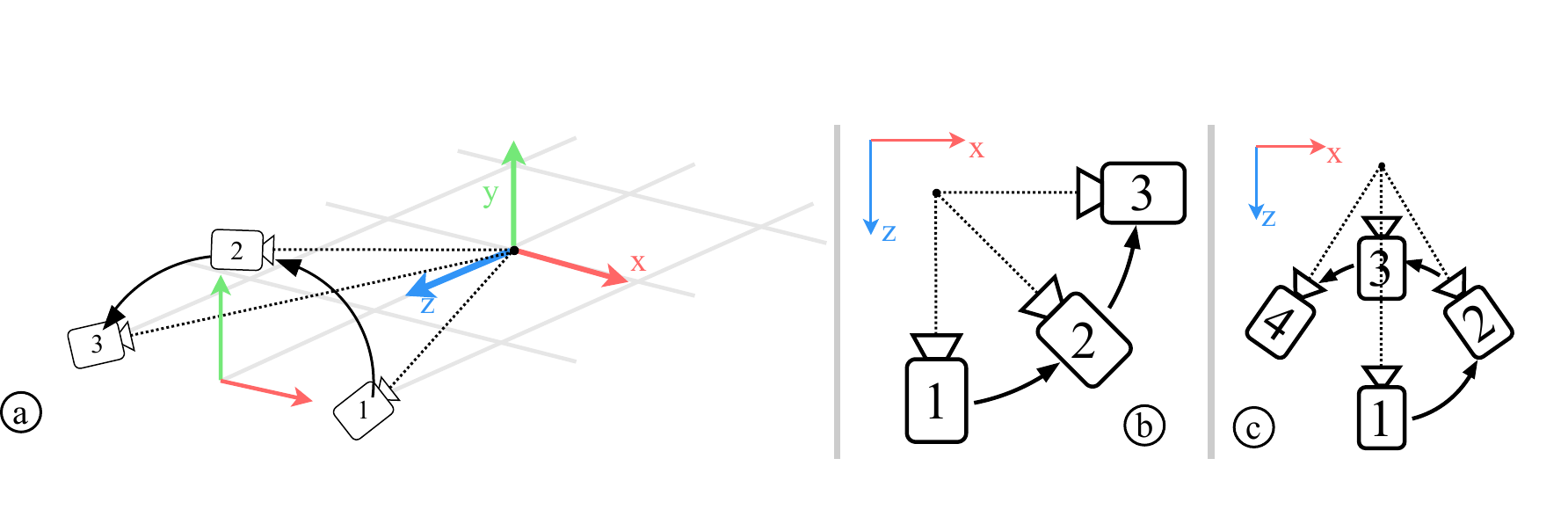}
       \vspace{-2.5em}
    \caption{A visualization of our custom trajectories: Hop (a), Orbit (b), and Spin (c).
    All cameras point towards a pivot in the scene, and the dotted lines represent the optical axes of the cameras. We use a coordinate space where $x$ is right, $y$ is up, and $z$ is backward.
    }
    \label{fig:custom-traj}
       \vspace{-10pt}
\end{figure}
\noindent\textbf{Image quality.}\ We evaluate the reconstruction performance of novel views using PSNR and LPIPS, across short-term and long-term generations.
Following previous work \cite{ren2022look}, we only consider test sequences where at least 200 frames are available, for RealEstate10K.
This choice ensures that there are ground truth images to evaluate against.
Starting with the first frame from the ground truth test sequences as our initial images, we generate 20 images of a sequence using 20 camera poses of the ground truth trajectory.
The camera poses are spaced ten frames apart with respect to the sequence's native frame rate, yielding a final camera pose that is 200 frames from the initial view.
Short-term evaluations are performed over the fifth generated image, and long-term evaluations are performed on the final generated image.
Quantitative results for RealEstate10K are provided in Tab.\ \ref{tab:recon-gt-all}.
Compared to the baselines, our method has the lowest reconstruction error in all cases.
We also evaluate LookOut \cite{ren2022look} without their additional post-processing step (LookOut-ne), and find that it yields slightly worse reconstruction results.

\begin{figure*}
    \vspace{-10pt}
    \centering
    \begin{subfigure}[b]{0.32\textwidth}
         \centering
         \includegraphics[width=\textwidth]{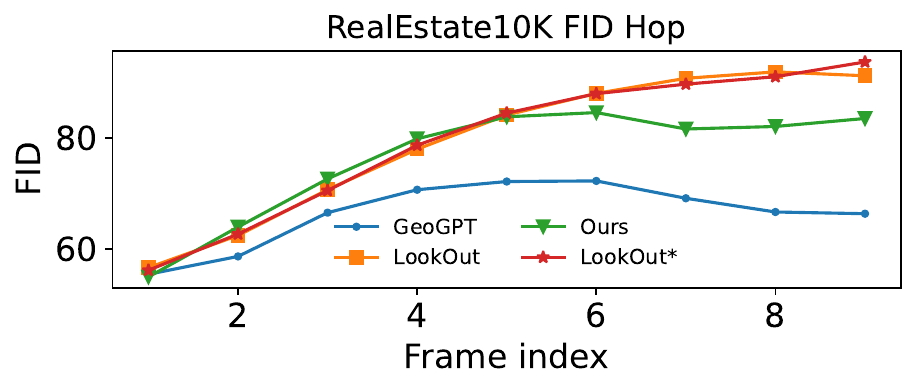}
         % \vspace{-20pt}
    \end{subfigure}
    \begin{subfigure}[b]{0.32\textwidth}
         \centering
         \includegraphics[width=\textwidth]{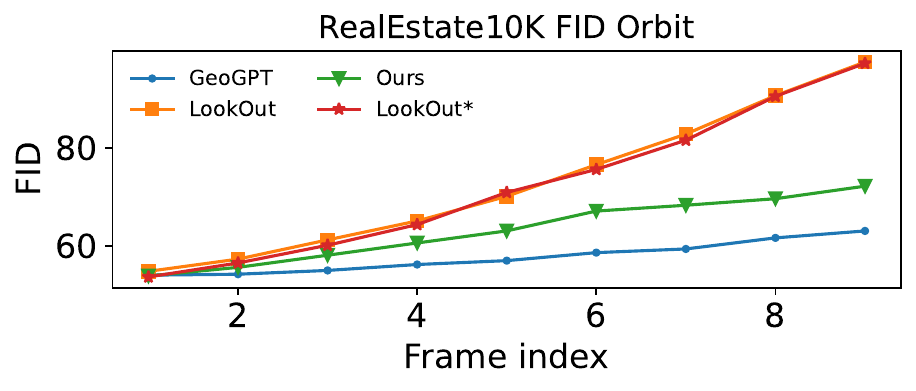}
         % \vspace{-20pt}
     \end{subfigure}
     \begin{subfigure}[b]{0.32\textwidth}
         \centering
         \includegraphics[width=\textwidth]{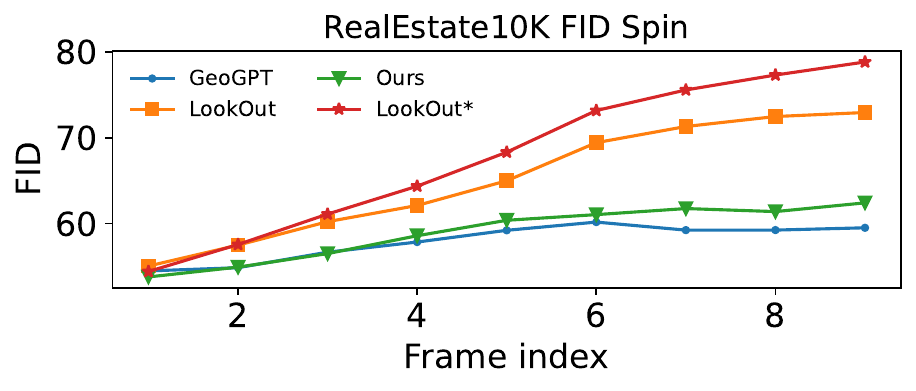}
         % \vspace{-20pt}
     \end{subfigure}
         \begin{subfigure}[b]{0.32\textwidth}
        \centering
        \includegraphics[width=\textwidth]{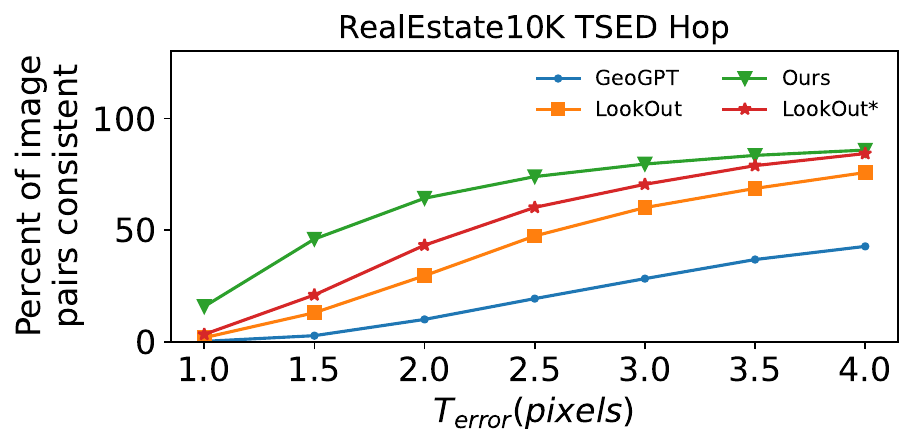}
        \vspace{-20pt}
    \end{subfigure}
    \begin{subfigure}[b]{0.32\textwidth}
        \centering
        \includegraphics[width=\textwidth]{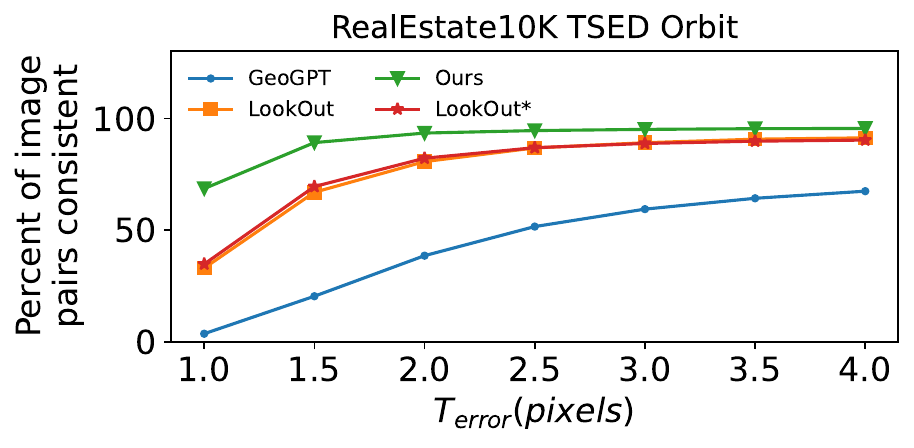}
        \vspace{-20pt}
        % \caption{Orbit}
    \end{subfigure}
    \begin{subfigure}[b]{0.32\textwidth}
        \centering
        \includegraphics[width=\textwidth]{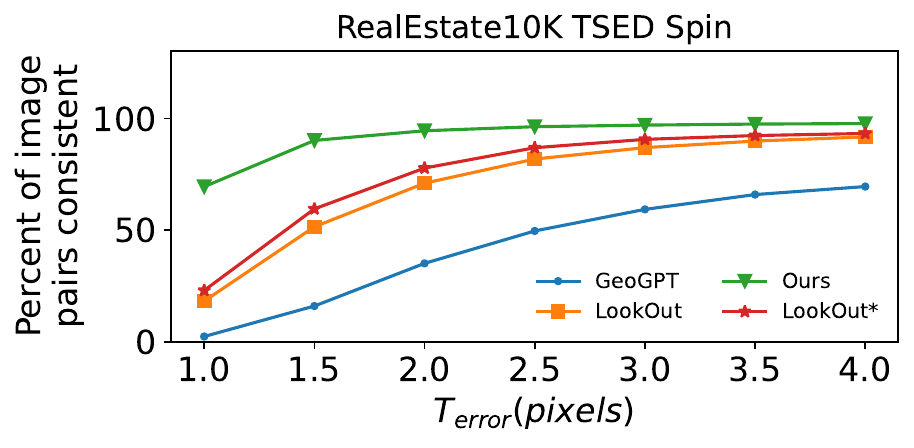}
        \vspace{-20pt}
        % \caption{Spin}
    \end{subfigure}
     % \vspace{-5pt}
     \caption{
     RealEstate10K FID (top), and TSED (bottom), on custom trajectories.
     Sequences are sampled using three novel trajectories designed to differ from the dominant modes in the dataset: Hop, Orbit, and Spin.
     \textit{LookOut*} is a version of LookOut without error accumulation post-processing.
     For TSED, we set $T_\text{matches} = 10$ while sweeping over a range of $T_\text{error}$ values.
     }
     \label{fig:fid-tsed-novel}
\end{figure*}
\begin{figure*}[t]
    \vspace{-10pt}
    \centering
    \begin{subfigure}[b]{0.3\textwidth}
         \centering
         \includegraphics[width=\textwidth]{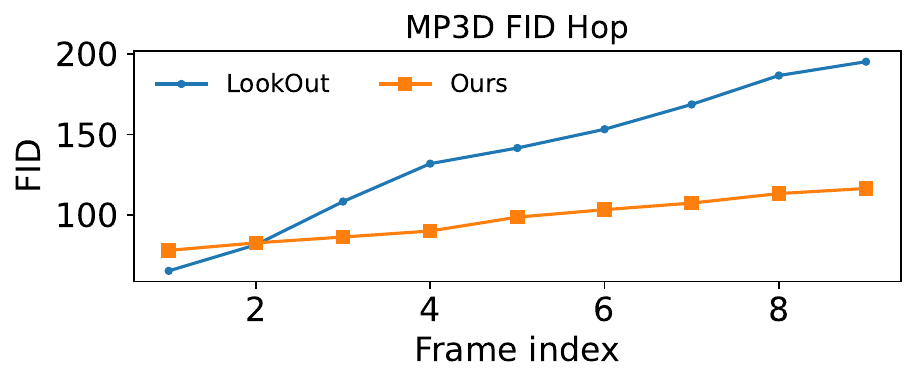}
         % \vspace{-20pt}
         % \caption{Hop}
    \end{subfigure}
    \begin{subfigure}[b]{0.3\textwidth}
         \centering
         \includegraphics[width=\textwidth]{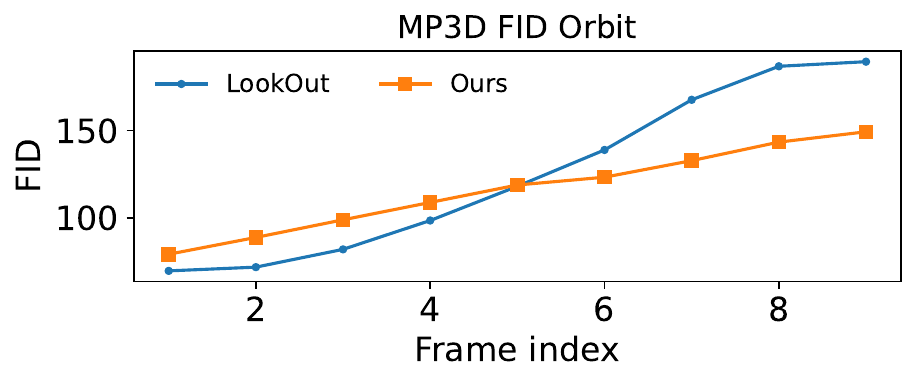}
         % \vspace{-20pt}
         % \caption{Orbit}
     \end{subfigure}
     \begin{subfigure}[b]{0.3\textwidth}
         \centering
         \includegraphics[width=\textwidth]{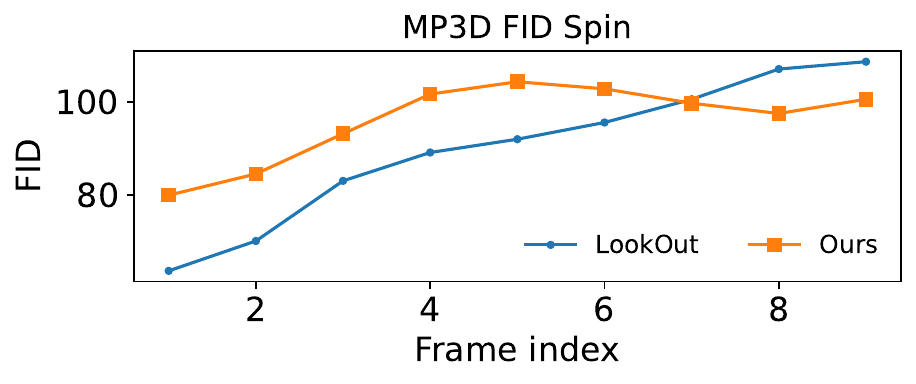}
         % \vspace{-20pt}
         % \caption{Spin}
     \end{subfigure}
      \begin{subfigure}[b]{0.3\textwidth}
         \centering
         \includegraphics[width=\textwidth]{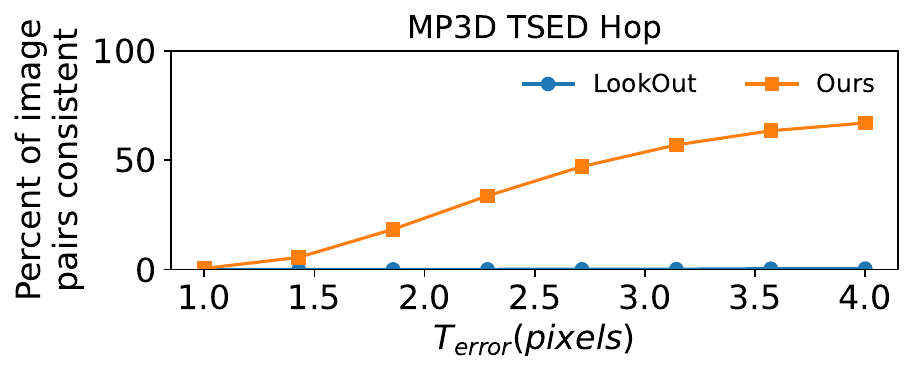}
         \vspace{-20pt}
         % \caption{Hop}
     \end{subfigure}
    \begin{subfigure}[b]{0.3\textwidth}
         \centering
         \includegraphics[width=\textwidth]{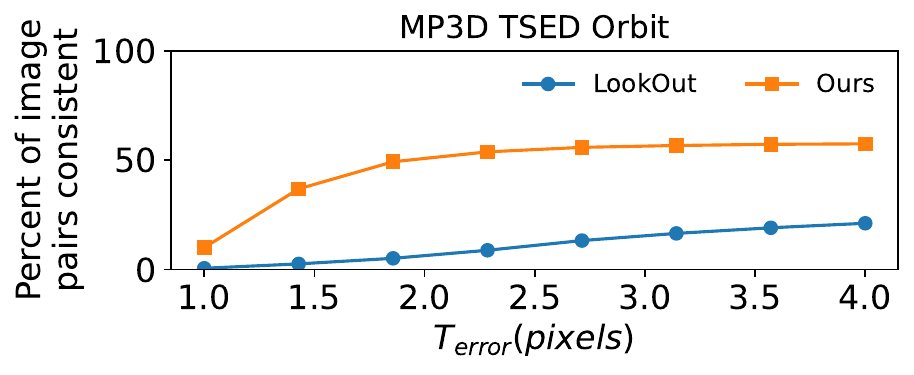}
         \vspace{-20pt}
         % \caption{Orbit}
     \end{subfigure}
     \begin{subfigure}[b]{0.3\textwidth}
         \centering
         \includegraphics[width=\textwidth]{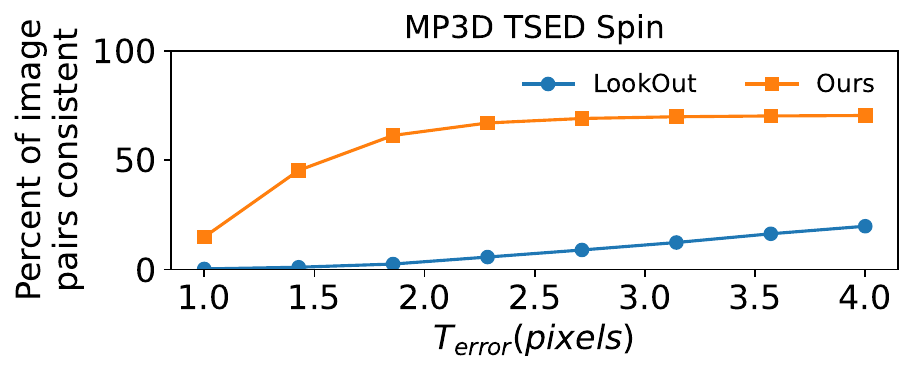}
         \vspace{-20pt}
         % \caption{Spin}
     \end{subfigure}
     % \vspace{-5pt}
     \caption{
     MP3D FID (top), and TSED (bottom), on custom trajectories.
     Evaluation on MP3D is performed using the same custom trajectories and TSED parameters as RealEstate10K.
     }
     \label{fig:fid-tsed-novel-mp3d}
     \vspace{-1em}
\end{figure*}
Similar to our full reference metric evaluation, we evaluate short-term and long-term quality with the no-reference metric FID.
To measure the generation image quality over time, we evaluate the FID between generated views at a specific time, and a fixed set of randomly selected views from the test set.
Tab.\ \ref{tab:FID-gt-all} presents quantitative results for RealEstate10K.
As seen from the table, all methods suffer from some level of error accumulation, and yield worse performance as the sequence length increases.
We find that LookOut produces images with significantly higher FID without the final error accumulation step.
For in-distribution trajectories, our method generates images with comparable quality as GeoGPT, and outperforms LookOut in terms of FID.
Notably, GeoGPT has the tendency to generate viewpoint-inconsistent images, where the semantics remain the same but the content changes.
This point is examined later using our viewpoint consistency metric.

In addition to RealEstate10K, we evaluate on MP3D with a similar setup, except the images in the sequence are neighboring frames since the rendered images from MP3D differ by larger camera motions.
We also provide reconstruction-based results for MP3D in Tab.\ \ref{tab:recon-gt-all}, and FID-based results in Tab.\ \ref{tab:FID-gt-all}, with LookOut as the baseline.
Quantitatively, we find the results with MP3D are similar to Realestate10K, where our method outperforms LookOut on all standard metrics for in-distribution trajectories.

\begin{figure*}
    \centering
    \includegraphics[width=0.99\textwidth]{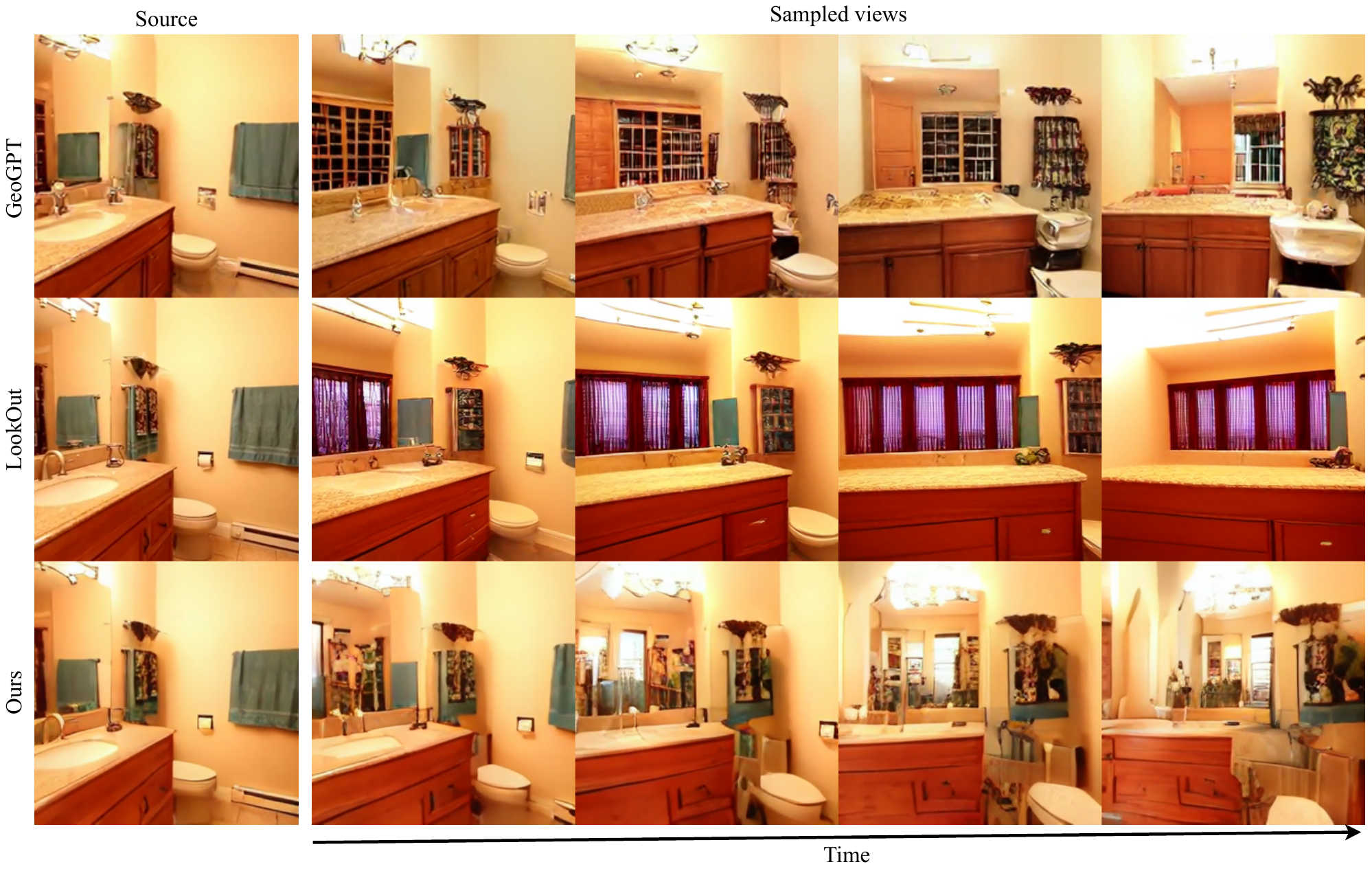}
    \vspace{-4pt}
    \caption{Samples from the \textit{Orbit} trajectory.
    Each row presents a generated image sequence from our method and the baselines, GeoGPT \cite{rombach2021geometry} and LookOut \cite{ren2022look}.
    The columns, are sampled views along the trajectory with the left most image being given.
    Notice both baselines give the impression of an orbiting camera motion, but parts of the visible scene in both views change between frames, \ie the cabinet under the sink. Images generated from LookOut tend to lose details in subsequent frames. Our method tends to maintain photometric consistency across the sequence.
    }
    \vspace{-1em}
    \label{fig:novel-samples}
\end{figure*}
\noindent\textbf{Consistency over long-term generations.}\ We evaluate the percent of consistent pairs of neighboring views out of 20 total pairs using our proposed metric, TSED (Sec.\ \ref{sec:tsed}). 
Quantitative results for RealEstate10K are shown in Fig.\ \ref{fig:cons-gt-both}, where we evaluate the consistency over a range of values for $T_\text{error}$, with $T_\text{matches} = 10$.
The average number of matches per pair on RealEstate10K is 33, 87, and 94, for GeoGPT, LookOut, and our method, respectively.
The lower consistency of GeoGPT is partly due to fewer matches per image pair.
Samples from LookOut have a comparable number of matches to our method, suggesting that the inconsistency is due to larger violations of the epipolar constraints.
Compared to the baselines, our method can generate better views with consistent appearances, and motion that respects epipolar constraints.
LookOut performs similarly on our consistency metric with and without error accumulation training.
We also compare LookOut and our method using TSED on MP3D, shown in Fig.\ \ref{fig:cons-gt-both}, and find that our method is more consistent on this dataset as well.

%======================

\subsection{Generation with Novel Trajectories}
Previous work limited evaluation to the ground truth trajectories in the RealEstate10K and MP3D datasets.
Consequently, given the biased nature of the trajectories, this may lead to overfitting.
Here, we explore the generalization capability of both our method and the baselines by evaluating on out-of-distribution trajectories.

As mentioned in Section \ref{sec:exp_setup}, the camera motions available in RealEstate10K, and MP3D are limited.
% Samples from out-of-distribution trajectories can be obtained by sampling autoregressively while ensuring that the relative poses of neighboring views are in-distribution.
We sample novel views over three manually defined trajectories distinct from those found in the training data: (i) a 90-degree orbit around the azimuth (Orbit), (ii) a vertical orbit along a semicircular path (Hop), and (iii) a translation along a circular path parallel to the ground plane (Spin).
These trajectories are illustrated in Fig.\ \ref{fig:custom-traj}.

As ground truth images are not available, we evaluate performance using the reference-free metrics, FID and TSED.
Quantitative results for RealEstate10K are summarized in Fig.\ \ref{fig:fid-tsed-novel}.
In terms of FID, our model's generation quality degrades faster than GeoGPT but slower than LookOut.
Qualitative results such as those shown in Fig.\ \ref{fig:novel-samples} suggest that when the baseline methods fail, they favour generating good-quality images, even though they may not be photometrically consistent with the other views.
The consistency of our generated sequences, evaluated using TSED, is higher than the baselines on all trajectories.
Between the three custom trajectories, \textit{Hop} is the most novel as it contains a vertical motion that is rare in RealEstate10K, while \textit{Spin} is the closest to the training trajectories, which contain many forward and backward motions.
Interestingly, LookOut without error accumulation performs better in the \textit{Hop} trajectory on TSED.
This suggests that the error accumulation post-processing may trade off generalization for higher image quality.
Overall, our method provides the best trade-off of photometric quality and consistency.

Fig.\ \ref{fig:fid-tsed-novel-mp3d} shows quantitative results on MP3D comparing LookOut, and our method.
LookOut is significantly less consistent than our method in terms of TSED, especially on \textit{Hop}.
Qualitative inspection reveals that LookOut generalizes poorly to our custom trajectories, and often does not generate images that respect the requested camera motion.
\section{Conclusion and Discussion}
We addressed the most challenging setting for NVS, \ie generative view extrapolation from a single image.
Our method exploits recent advancements in diffusion-based generative models to sample multiple consistent novel views. 
Empirically, we presented a finer-grained evaluation of the task compared to previous studies.
In particular, reported results of previous work focus on generated image quality of each image but ignore geometric consistency.
Here, we introduced a new metric based on epipolar geometry, which directly evaluates geometric consistency of generated views independent of image quality.
Based on both new and standard metrics, we showed that our method generates images that are more consistent than current methods, while maintaining high image quality.
Further, on camera trajectories that are atypical of the training data, we showed that our method generates images that are more consistent than the baselines.
\\
\\
\noindent\textbf{Acknowledgements.}\ This work was funded in part by the Canada First Research Excellence Fund (CFREF) for the Vision: Science to Applications (VISTA) program, the NSERC Discovery Grant program, the NSERC Canada Graduate Scholarships – Doctoral program, and the Vector Institute for AI.

%%%%%%%%% REFERENCES
% \clearpage
{\small
\bibliographystyle{ieee_fullname}
\bibliography{refs_short,refs}
}

\clearpage
\appendix
\onecolumn
% change table and figure header to include section letter
\renewcommand\thetable{\thesection.\arabic{table}}
\renewcommand\thefigure{\thesection.\arabic{figure}}

\begin{center}
    \Large
    Appendix
\end{center}

\section{Architecture Details}
\label{sec:arch}
In this section, we provide additional details of our model described in Section 3.2 of the main paper.
Our model is based on \textit{Noise Conditional Score Network++} (NCSN++) \cite{song2020score}.
An overview of the main backbone is provided in Tables \ref{tab:arch-encoder} and \ref{tab:arch-decoder}.
Two streams of the backbone are used to process the conditioning and generated image.
We modify the original architecture by adding cross-attention layers throughout the backbone, which attend to features in the opposite stream.
The residual blocks are based on the residual blocks used in BigGAN \cite{brock2018large}.
Upsampling and downsampling is also performed in the network using BigGAN residual blocks \cite{brock2018large}.
Inputs to the backbone encoder are provided at various layers using a multi-scale pyramid.
Outputs of the network are accumulated from multiple layers of the decoder using a multi-scale residual pyramid.
Specific implementation details can be found in the code release: \webpage.
\begin{table}[h]
    \scriptsize
    \parbox{.49\linewidth}{
    \centering
    \begin{tabular}{c|c|c|c}
        Layer         &      Output size      & Additional inputs & Additional outputs \\
        \hline
        Input image   &  $4\times32\times32$  &                   & Skip 0,In Pyramid  \\
        \hline
        ResBlock      &  $256\times32\times32$ &  Time emb.        &                    \\
        Spatial Attn. &  $256\times32\times32$ &                   &                    \\
        Cross Attn.   &  $256\times32\times32$ &  Cross, Rays      & Skip 1, Cross      \\
        ResBlock      &  $256\times32\times32$ &  Time emb.        &                    \\
        Spatial Attn. &  $256\times32\times32$ &                   &                    \\
        Cross Attn.   &  $256\times32\times32$ &  Cross, Rays      & Skip 2, Cross      \\
        ResBlockDown  &  $256\times16\times16$ &  Time emb.        &                    \\
        Combiner      &  $256\times16\times16$ &  In Pyramid 1     & Skip 3             \\
        \hline
        ResBlock      &  $256\times16\times16$ &  Time emb.        &                    \\
        Spatial Attn. &  $256\times16\times16$ &                   &                    \\
        Cross Attn.   &  $256\times16\times16$ &  Cross, Rays      & Skip 4, Cross      \\
        ResBlock      &  $256\times16\times16$ &  Time emb.        &                    \\
        Spatial Attn. &  $256\times16\times16$ &                   &                    \\
        Cross Attn.   &  $256\times16\times16$ &  Cross, Rays      & Skip 5, Cross      \\
        ResBlockDown  &   $256\times8\times8$  &  Time emb.        &                    \\
        Combiner      &   $256\times8\times8$  &  In Pyramid 2     & Skip 6             \\
        \hline
        ResBlock      &   $256\times8\times8$  &  Time emb.        &                    \\
        Spatial Attn. &   $256\times8\times8$  &                   &                    \\
        Cross Attn.   &   $256\times8\times8$  &  Cross, Rays      & Skip 7, Cross      \\
        ResBlock      &   $256\times8\times8$  &  Time emb.        &                    \\
        Spatial Attn. &   $256\times8\times8$  &                   &                    \\
        Cross Attn.   &   $256\times8\times8$  &  Cross, Rays      & Skip 8, Cross      \\
        ResBlockDown  &   $256\times4\times4$  &  Time emb.        &                    \\
        Combiner      &   $256\times4\times4$  &  In Pyramid 3     & Skip 9             \\
        \hline
        ResBlock      &   $256\times4\times4$  &  Time emb.        &                    \\
        Spatial Attn. &   $256\times4\times4$  &                   &                    \\
        Cross Attn.   &   $256\times4\times4$  &  Cross, Rays      & Skip 10, Cross     \\
        ResBlock      &   $256\times4\times4$  &  Time emb.        &                    \\
        Spatial Attn. &   $256\times4\times4$  &                   &                    \\
        Cross Attn.   &   $256\times4\times4$  &  Cross, Rays      & Skip 11, Cross     \\
        \hline
        \hline
        ResBlock      &    $256\times4\times4$  &  Time emb.        &                    \\
        Spatial Attn. &    $256\times4\times4$  &                   &                    \\
        ResBlock      &    $256\times4\times4$  &  Time emb.        &                    \\
    \end{tabular}
    \caption{NSCN++ U-Net backbone encoder.
    ResBlocks are BigGAN \cite{brock2018large} style residual blocks, ResBlocksDown layers are the same, but configured with a downsampling option.
    Time emb.\ is the time information provided for the diffusion model.
    Skip inputs are skip connections that go to the decoder.
    Rays are the camera ray conditioning, and Cross is a cross-attention connection to the other stream.}
    \label{tab:arch-encoder}
    }
    \hfill
    \parbox{.49\linewidth}{
    \centering
    \begin{tabular}{c|c|c|c}
        Layer             &      Output size      & Additional inputs   & Additional outputs \\
        \hline
        Encoder input     &  $256\times4\times4$  &                     &                    \\
        \hline
        ResBlock          &  $256\times4\times4$  &  Time emb., Skip 11 &                    \\
        ResBlock          &  $256\times4\times4$  &  Time emb., Skip 10 &                    \\
        ResBlock          &  $256\times4\times4$  &  Time emb., Skip 9  &                    \\
        Spatial Attn.     &  $256\times4\times4$  &                     &                    \\
        Cross Attn.       &  $256\times4\times4$  &  Cross, Rays        &  Cross             \\
        Conv$3\times3$    &  $256\times4\times4$  &  Out Pyramid 1      &                    \\
        ResBlockUp          &  $256\times8\times8$  &  Time emb.          &                    \\
        \hline  
        ResBlock          &  $256\times8\times8$  &  Time emb., Skip 8  &                    \\
        ResBlock          &  $256\times8\times8$  &  Time emb., Skip 7  &                    \\
        ResBlock          &  $256\times8\times8$  &  Time emb., Skip 6  &                    \\
        Spatial Attn.     &  $256\times8\times8$  &                     &                    \\
        Cross Attn.       &  $256\times8\times8$  &  Cross, Rays        &  Cross             \\
        Conv$3\times3$    &  $256\times8\times8$  &  Out Pyramid 2      &                    \\
        ResBlockUp          & $256\times16\times16$ &  Time emb.          &                    \\
        \hline
        ResBlock          & $256\times16\times16$ &  Time emb., Skip 5  &                    \\
        ResBlock          & $256\times16\times16$ &  Time emb., Skip 4  &                    \\
        ResBlock          & $256\times16\times16$ &  Time emb., Skip 3  &                    \\
        Spatial Attn.     & $256\times16\times16$ &                     &                    \\
        Cross Attn.       & $256\times16\times16$ &  Cross, Rays        &  Cross             \\
        Conv$3\times3$    & $256\times16\times16$ &  Out Pyramid 3      &                    \\
        ResBlockUp          & $256\times32\times32$ &  Time emb.          &                    \\
        \hline
        ResBlock          & $256\times32\times32$ &  Time emb., Skip 2  &                    \\
        ResBlock          & $256\times32\times32$ &  Time emb., Skip 1  &                    \\
        ResBlock          & $256\times32\times32$ &  Time emb., Skip 0  &                    \\
        Spatial Attn.     & $256\times32\times32$ &                     &                    \\
        Cross Attn.       & $256\times32\times32$ &  Cross, Rays        &  Cross             \\
        Conv$3\times3$    & $256\times32\times32$ &  Out Pyramid 4      &                    \\
    \end{tabular}
    \caption{NSCN++ U-Net backbone decoder.
    ResBlocks are BigGAN \cite{brock2018large} style residual blocks, ResBlocksUp layers are the same, but configured with an upsampling option. Time emb.\ is the time information provided for the diffusion model.
    Skip inputs are skip connections coming from the encoder.
    Rays are the camera ray conditioning, and Cross is the cross-attention connection to the other stream.}
    \label{tab:arch-decoder}
    }
\end{table}

\section{TSED Sensitivity Analysis.}
\label{sec:sensitivity}
\begin{figure*}[t]
    \centering
    \begin{subfigure}[b]{0.49\textwidth}
         \centering
         \includegraphics[width=\linewidth]{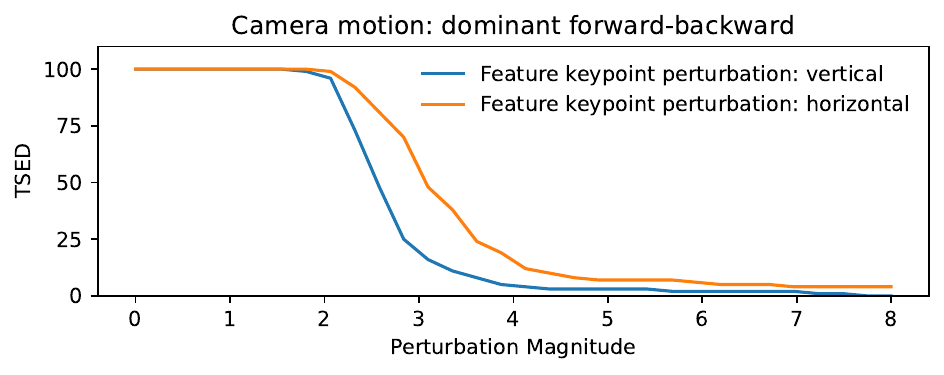}
    \end{subfigure}
    \begin{subfigure}[b]{0.49\textwidth}
         \centering
         \includegraphics[width=\linewidth]{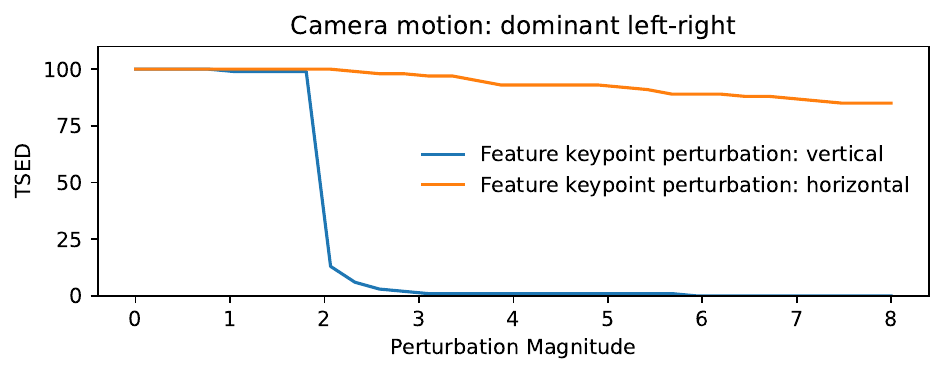}
    \end{subfigure}
    \begin{subfigure}[b]{0.5\textwidth}
         \centering
         \includegraphics[width=\linewidth]{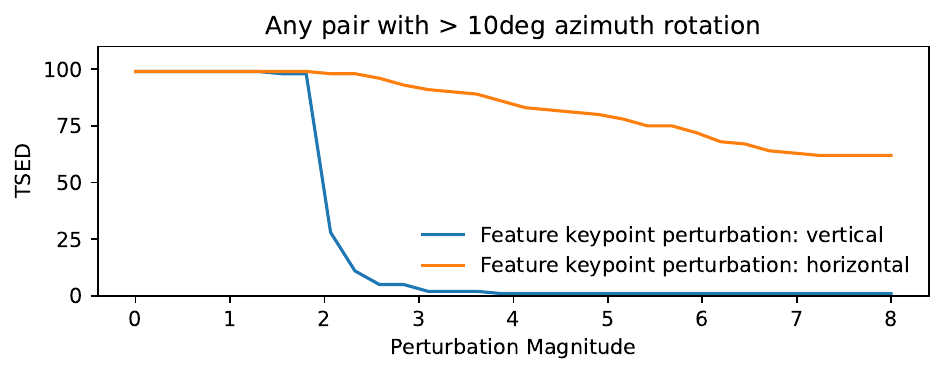}
    \end{subfigure}
    \vspace{-8pt}
    \caption{
    TSED sensitivity analysis for image pairs with different dominant camera motions using $T_\text{error}=2$.
    TSED scores are plotted for perturbations to the 2D correspondence locations with constant magnitude along horizontal and vertical directions.
    Camera motion determines the orientations of the epipolar lines, which can make the metric insensitive in some cases when many epipolar lines share the same orientation.
    }
    \label{fig:tsed-sensitivity}
    \vspace{-10pt}
\end{figure*}

A drawback to using epipolar geometry to measure consistency between correspondences and the camera poses is the potential for TSED to be insensitive to positional errors in the correspondences along epipolar lines.
We empirically analyse the sensitivity of TSED on ground truth image pairs from RealEstate10K \cite{realestate46965} under three classes of camera motion: dominant forward-backward motions, dominant left-right motions, and motion that contains more than ten degrees of azimuth rotation.
Using $T_\text{error}=2$, we compute TSED over 100 random image pairs in each class while adding perturbations to the 2D positions of the correspondences in each view by a constant magnitude along horizontal and vertical directions.
In the ideal case when TSED is maximally sensitive, it should show a sharp reduction when the perturbations have a magnitude of $T_\text{error}$ or greater.
Results from our sensitivity analysis are shown in Fig.\ \ref{fig:tsed-sensitivity}.
As expected, TSED is least sensitive to horizontal perturbations for when there are left-right camera motions since most of the epipolar lines are horizontal.
For image-pairs with greater than 10 degrees of azimuth rotation, there are fewer horizontal epipolar lines, and TSED is more sensitive to horizontal perturbations than with dominant left-right motion.
The results also show that TSED is most sensitive for forward-backward motions since the epipolar lines have a variety of orientations.

\section{Stochastic Conditioning on RealEstate10K}
\label{sec:sc}
\begin{figure}[h]
    \centering
    \begin{subfigure}[b]{0.33\textwidth}
         \centering
         \includegraphics[width=\textwidth]{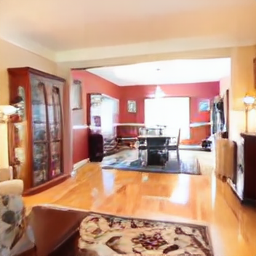}
         \caption{Source image.}
     \end{subfigure}
     \begin{subfigure}[b]{0.33\textwidth}
         \centering
         \includegraphics[width=\textwidth]{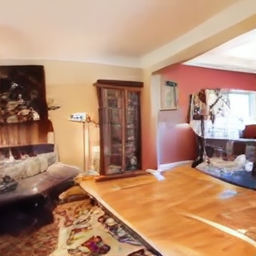}
         \caption{Frame 5 of Markov sampling.}
     \end{subfigure}
     \begin{subfigure}[b]{0.33\textwidth}
         \centering
         \includegraphics[width=\textwidth]{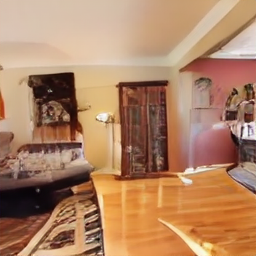}
         \caption{Frame 7 of Markov sampling.}
     \end{subfigure}
     \hspace*{\fill}
     \begin{subfigure}[b]{0.33\textwidth}
         \centering
         \includegraphics[width=\textwidth]{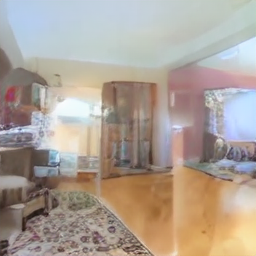}
         \caption{Frame 5 with stochastic conditioning.}
     \end{subfigure}
     \begin{subfigure}[b]{0.33\textwidth}
         \centering
         \includegraphics[width=\textwidth]{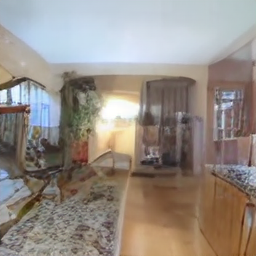}
         \caption{Frame 7 with stochastic conditioning.}
     \end{subfigure}
    \caption{Comparison of generation using a Markov dependency vs stochastic conditioning with the previous two frames as input.
    Both methods were generated using the same trajectory and source image.
    Notice the reduction of quality when stochastic conditioning is applied.}
    \label{fig:stochastic-conditioning}
\end{figure}

Previous work \cite{watson2022novel} proposed a heuristic for extending a novel view diffusion model to use an arbitrary number of source views, called \textit{stochastic conditioning}.
Given $m$ possible source views, each iteration of the diffusion sampling process is modified to be randomly conditioned on one of the $m$ views.
Results using stochastic conditioning on CLEVR \cite{johnson2017clevr} are provided in the main paper in Section 4.2.
Previous work \cite{watson2022novel} used stochastic conditioning to condition on all previous frames.
We also apply this heuristic for generating sets of views on RealEstate10K \cite{realestate46965}, but we conditioned on up to two of the previous frames.
Qualitative results shown in Figure \ref{fig:stochastic-conditioning} exhibit a significant reduction in quality, and contain noticeable artifacts.
As a consequence, we did not include results based on stochastic conditioning with our method.

\section{Additional Qualitative Results}
\label{sec:more-qual}
Additional qualitative results are provided with an interactive viewer on our project page, \webpage, under the \textbf{RealEstate10K Qualitative Results - Out-of-Distribution Trajectories} and \textbf{RealEstate10K Qualitative Results - In-Distribution Trajectories} sections. The viewer allows the images along a trajectory to be explored for multiple scenes, and sampling instances.
Due to the stochastic nature of our model and the baselines, different plausible extrapolations of the scene are shown in the different instances of sampling.
Additional qualitative results for Matterport3D \cite{chang2017matterport3d} are also available on our project page, \webpage, under the \textbf{Matterport3D Qualitative Results - Out-of-Distribution Trajectories} and \textbf{Matterport3D Qualitative Results - In-Distribution Trajectories} sections.

\clearpage
\section{Additional Results with TSED}
\label{sec:more-tsed}
\begin{figure}[H]
    \centering
    \begin{subfigure}[b]{0.24\textwidth}
         \centering
         \includegraphics[width=\textwidth]{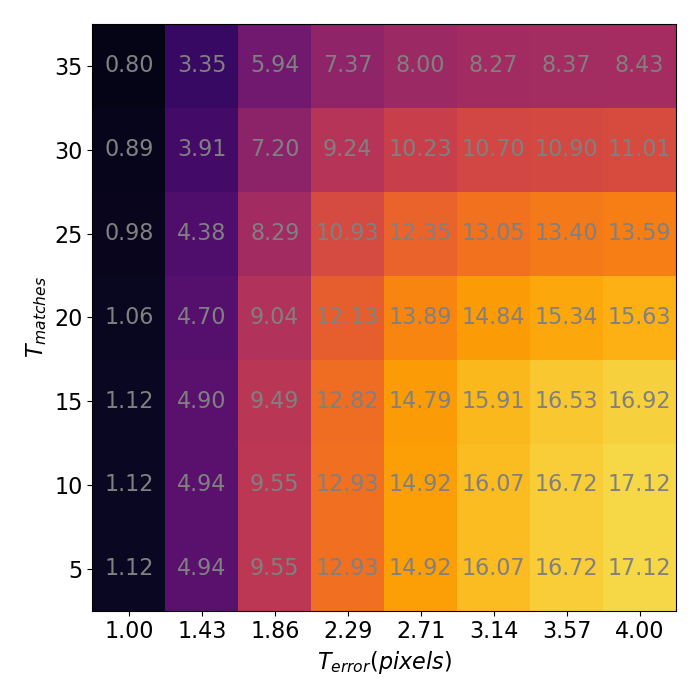}
         \caption{GeoGPT}
     \end{subfigure}
     \begin{subfigure}[b]{0.24\textwidth}
         \centering
         \includegraphics[width=\textwidth]{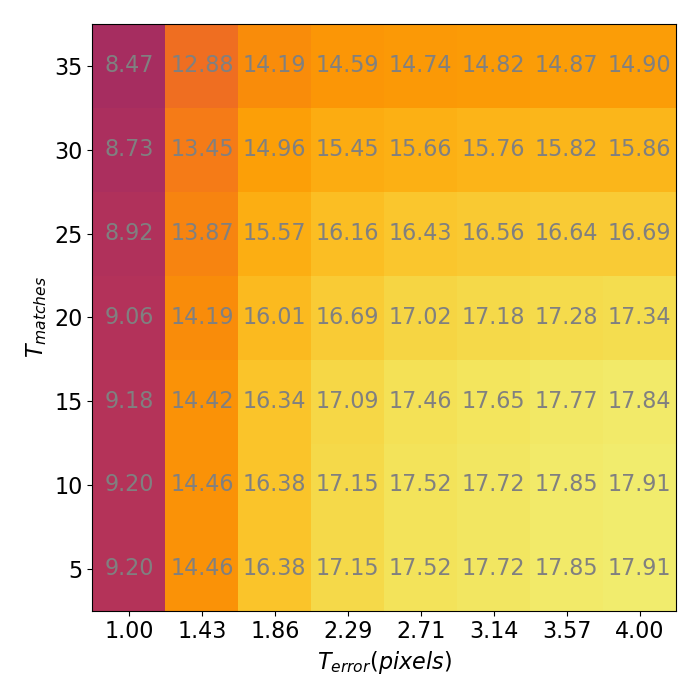}
         \caption{Lookout-noerror}
     \end{subfigure}
     \begin{subfigure}[b]{0.24\textwidth}
         \centering
         \includegraphics[width=\textwidth]{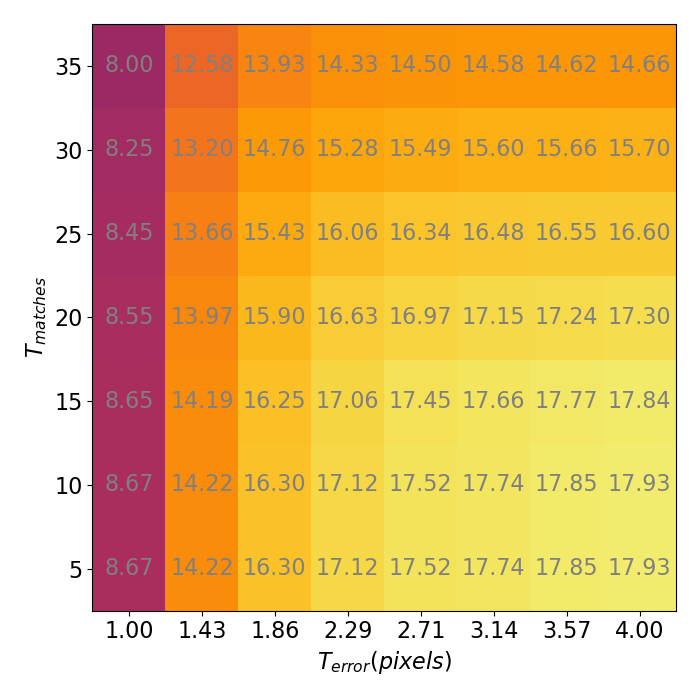}
         \caption{Lookout}
     \end{subfigure}
     \begin{subfigure}[b]{0.24\textwidth}
         \centering
         \includegraphics[width=\textwidth]{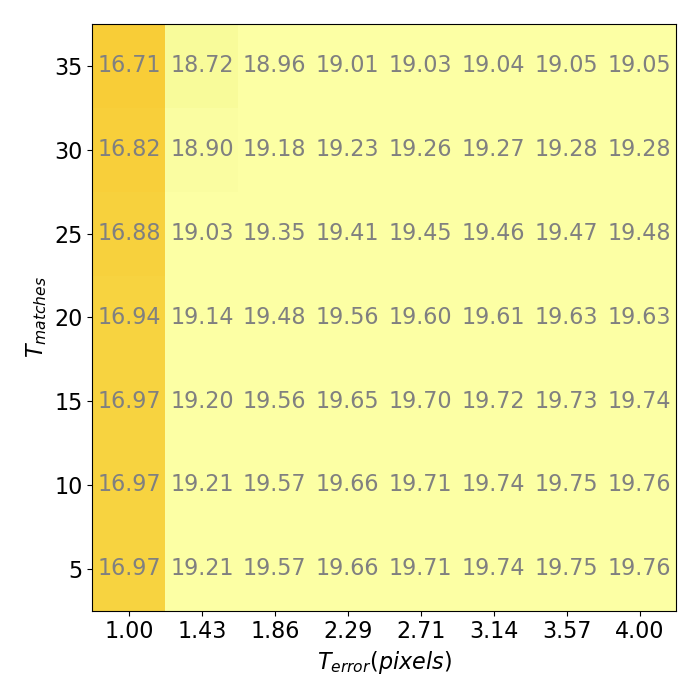}
         \caption{Ours}
     \end{subfigure}
    \caption{TSED computed using images generated over in-distribution trajectories. We sweep over a range of values for $T_\text{matches}$ and $T_\text{error}$.
    The values are provided as the average number of consistent pairs per sequence out of 20.}
    \label{fig:tsed_sweep_real}
\end{figure}

\begin{figure}[H]
    \centering
    \begin{subfigure}[b]{0.24\textwidth}
         \centering
         \includegraphics[width=\textwidth]{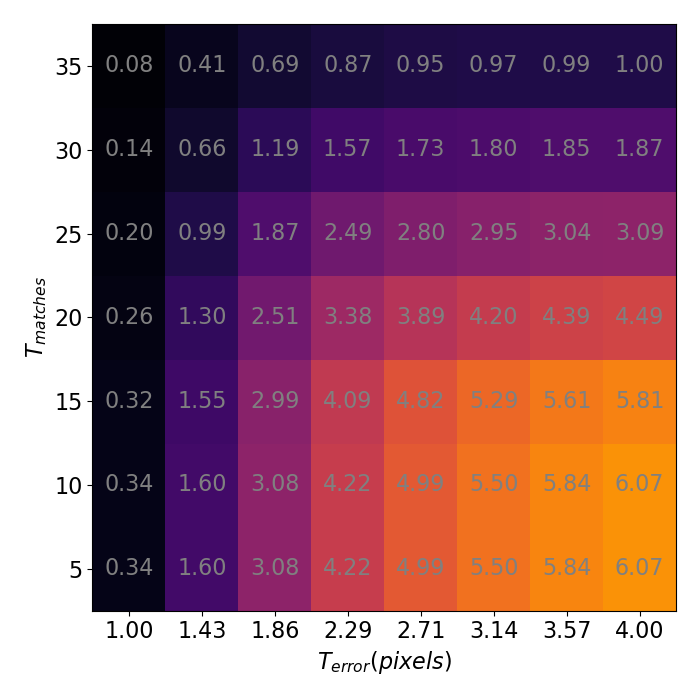}
         \caption{GeoGPT}
     \end{subfigure}
     \begin{subfigure}[b]{0.24\textwidth}
         \centering
         \includegraphics[width=\textwidth]{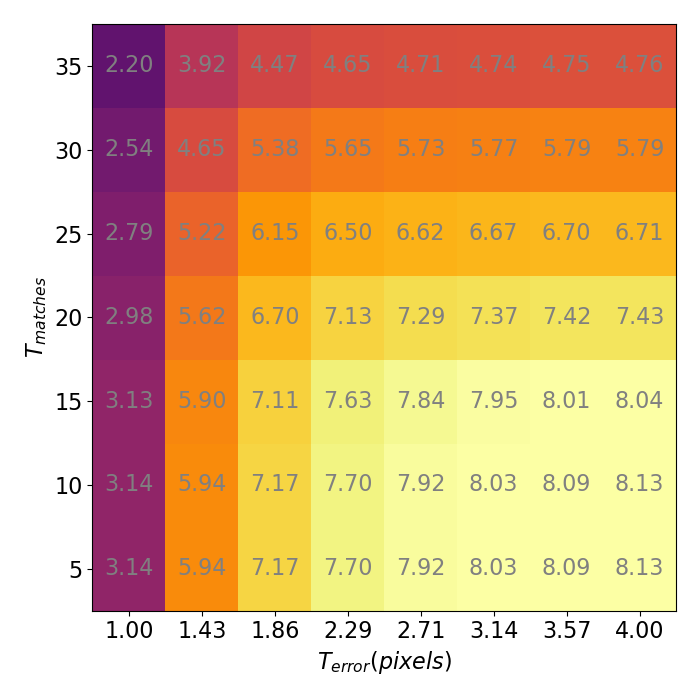}
         \caption{Lookout-noerror}
     \end{subfigure}
     \begin{subfigure}[b]{0.24\textwidth}
         \centering
         \includegraphics[width=\textwidth]{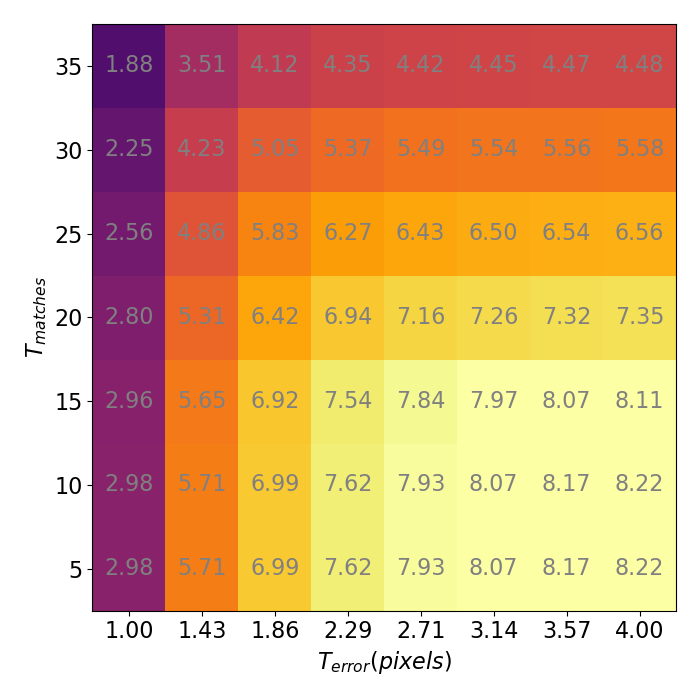}
         \caption{Lookout}
     \end{subfigure}
     \begin{subfigure}[b]{0.24\textwidth}
         \centering
         \includegraphics[width=\textwidth]{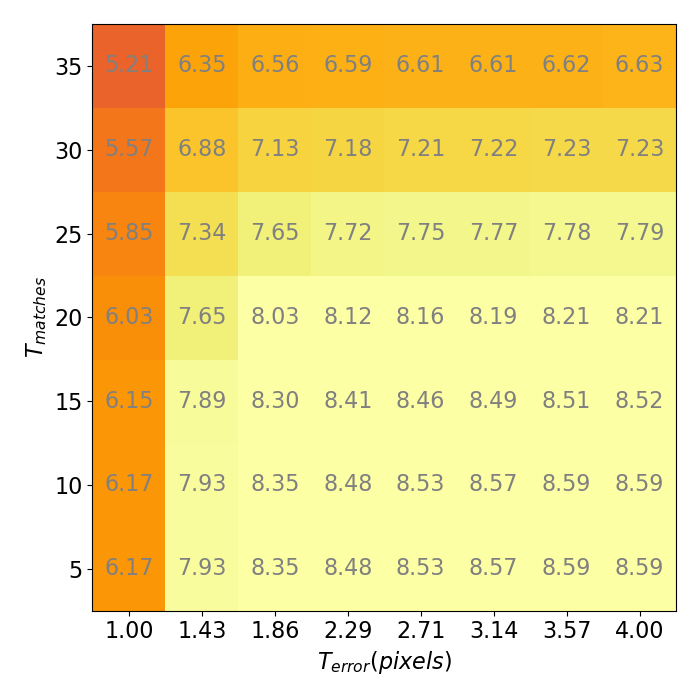}
         \caption{Ours}
     \end{subfigure}
    \caption{TSED computed using images generated over \textit{orbit} trajectory. We sweep over a range of values for $T_\text{matches}$ and $T_\text{error}$.
    The values are provided as the average number of consistent pairs per sequence out of 9.}
    \label{fig:tsed_sweep_orbit}
\end{figure}

\begin{figure}[H]
    \centering
    \begin{subfigure}[b]{0.24\textwidth}
         \centering
         \includegraphics[width=\textwidth]{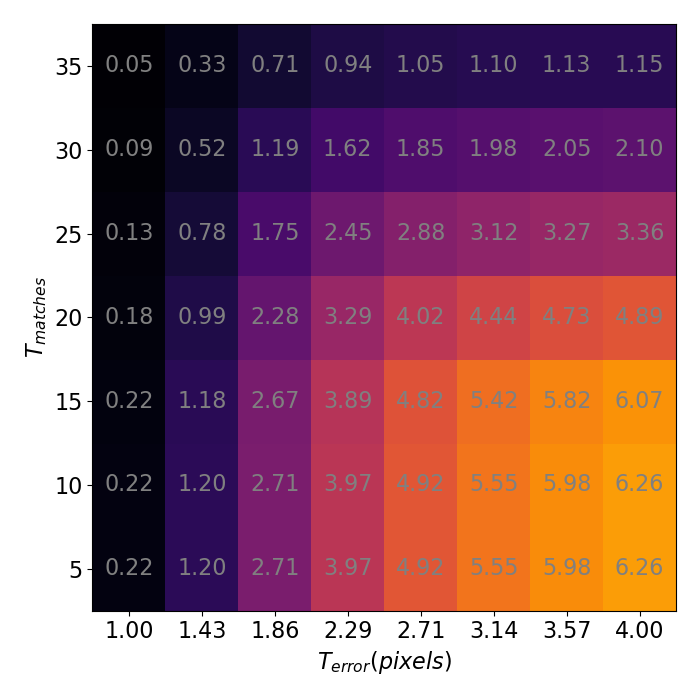}
         \caption{GeoGPT}
     \end{subfigure}
     \begin{subfigure}[b]{0.24\textwidth}
         \centering
         \includegraphics[width=\textwidth]{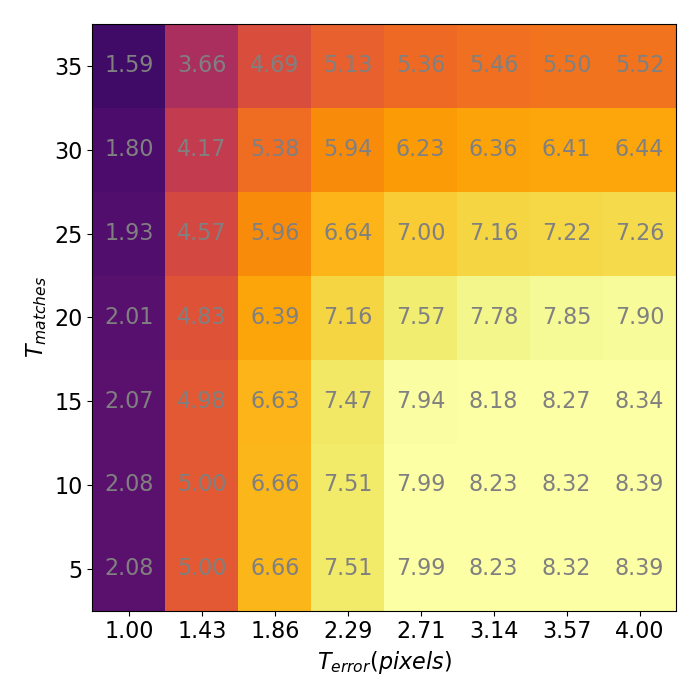}
         \caption{Lookout-noerror}
     \end{subfigure}
     \begin{subfigure}[b]{0.24\textwidth}
         \centering
         \includegraphics[width=\textwidth]{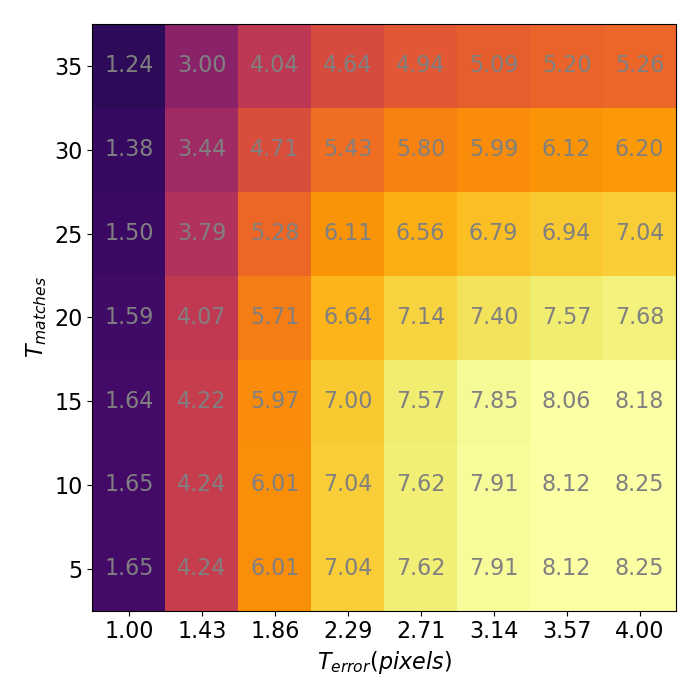}
         \caption{Lookout}
     \end{subfigure}
     \begin{subfigure}[b]{0.24\textwidth}
         \centering
         \includegraphics[width=\textwidth]{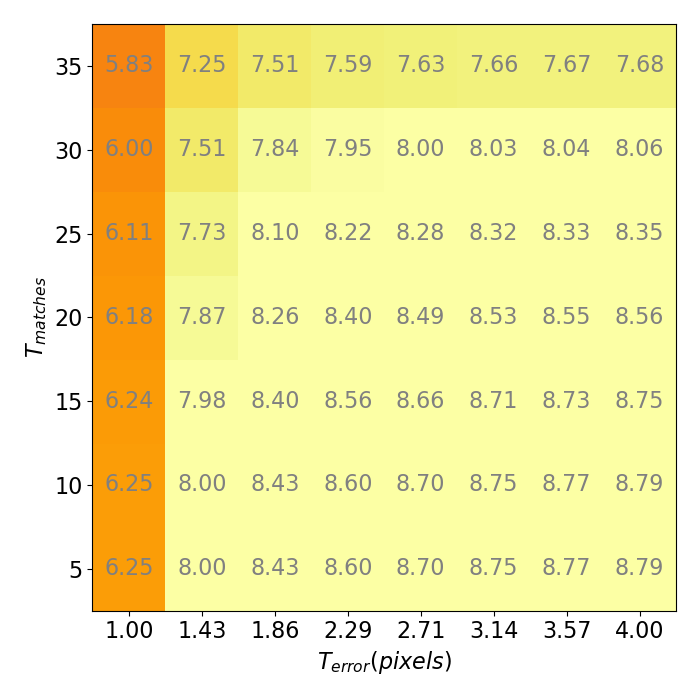}
         \caption{Ours}
     \end{subfigure}
    \caption{TSED computed using images generated over \textit{spin} trajectory. We sweep over a range of values for $T_\text{matches}$ and $T_\text{error}$.
    The values are provided as the average number of consistent pairs per sequence out of 9.}
    \label{fig:tsed_sweep_spin}
\end{figure}

\begin{figure}[H]
    \centering
    \begin{subfigure}[b]{0.24\textwidth}
         \centering
         \includegraphics[width=\textwidth]{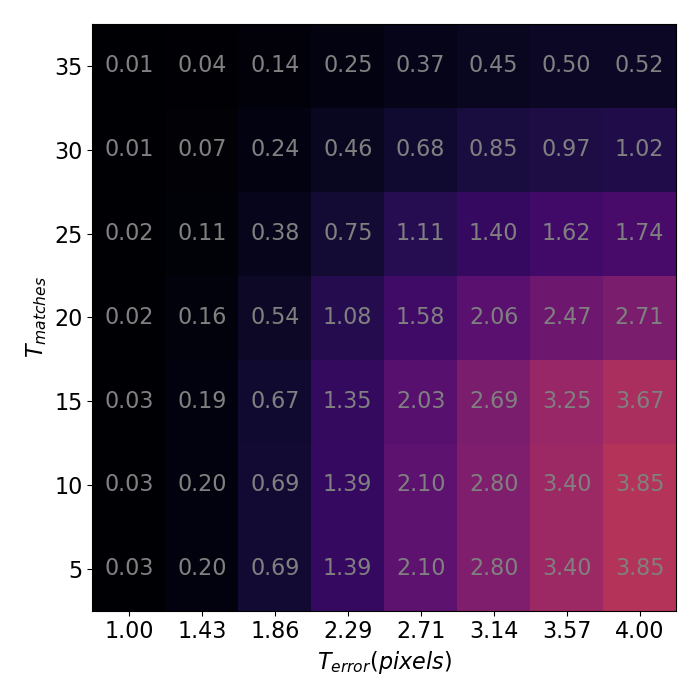}
         \caption{GeoGPT}
     \end{subfigure}
     \begin{subfigure}[b]{0.24\textwidth}
         \centering
         \includegraphics[width=\textwidth]{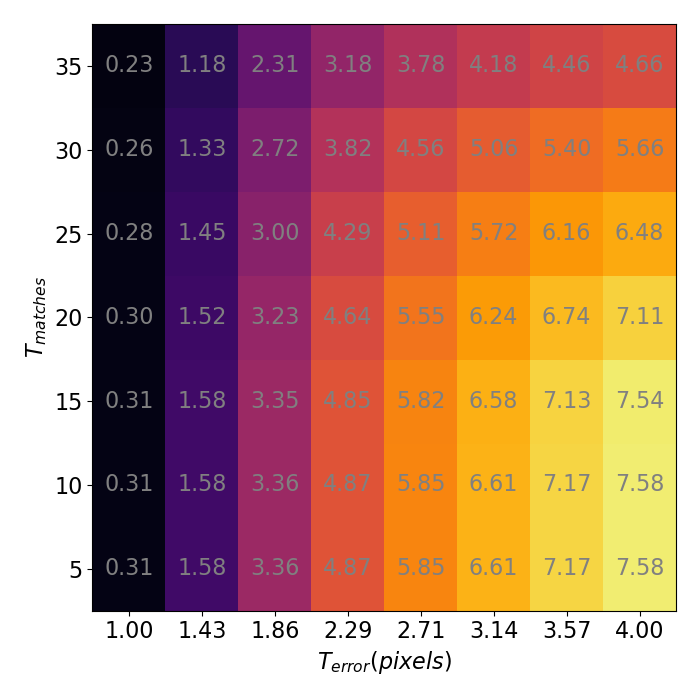}
         \caption{Lookout-noerror}
     \end{subfigure}
     \begin{subfigure}[b]{0.24\textwidth}
         \centering
         \includegraphics[width=\textwidth]{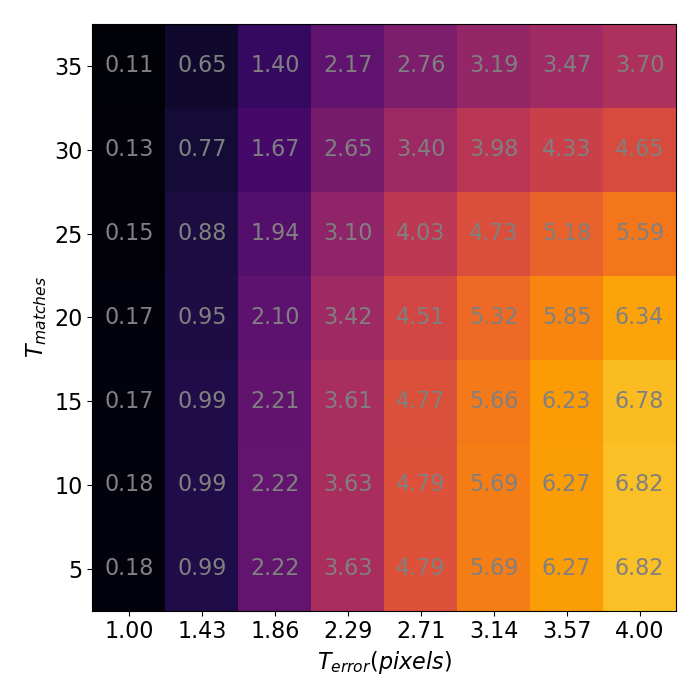}
         \caption{Lookout}
     \end{subfigure}
     \begin{subfigure}[b]{0.24\textwidth}
         \centering
         \includegraphics[width=\textwidth]{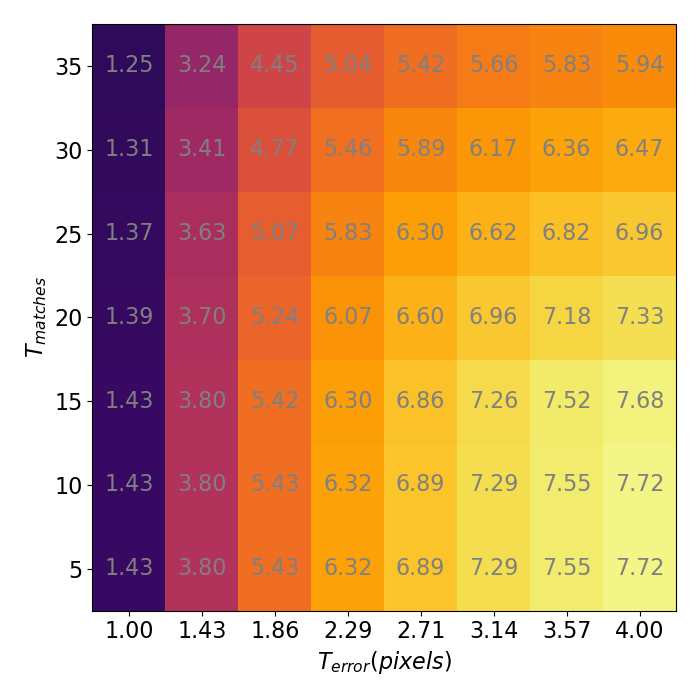}
         \caption{Ours}
     \end{subfigure}
    \caption{TSED computed using images generated over our \textit{hop} trajectory. We sweep over a range of values for $T_\text{matches}$ and $T_\text{error}$.
    The values are provided as the average number of consistent pairs per sequence out of 9.}
    \label{fig:tsed_sweep_hop}
\end{figure}

We provide additional quantitative results using TSED in Figures \ref{fig:tsed_sweep_real}, \ref{fig:tsed_sweep_orbit}, \ref{fig:tsed_sweep_spin}, and \ref{fig:tsed_sweep_hop} for images generated using in-distribution trajectories, and the orbit, spin, hop out-of-distribution trajectories, respectively.
We sweep across a range of values for both $T_\textbf{error}$ and $T_\textbf{matches}$.
Pairs of images with less than $T_\textbf{matches}$ SIFT \cite{lowe1999object} matches, or a median SED \cite{hartley2003multiple} lower than $T_\textbf{error}$, are considered not consistent.
In all trajectory types, GeoGPT \cite{rombach2021geometry} is the most affected by $T_\textbf{matches}$ due to a lack of photometric consistency, which leads to a low number of SIFT correspondences.
The TSED for both variants of Lookout \cite{ren2022look} do not vary as severely as GeoGPT with respect to $T_\textbf{matches}$.
Image pairs generated with our method tend to yield more SIFT matches, and are mainly affected by $T_\textbf{error}$.
The quantitative TSED results in the main paper were evaluated at $T_\textbf{matches}=10$, but these extended results show that our method yields higher TSED scores, remains consistent over a range of $T_\textbf{matches}$ values, in all cases.


To provide a better intuition on how symmetric epipolar distance (SED) \cite{hartley2003multiple} provides a measure of consistency, we provide an interactive demo on our project page, \webpage, under the \textbf{Visualization of SED} section.
The demo visualizes how SED varies in response to the positions of two correspondences in a pair of views with known relative camera geometry.
Each point creates an epipolar line on the opposite image, and the minimal distance line between a point and a line on the same image is shown.

\section{Limitations of Autoregressive Sampling}
\label{sec:limits}
\begin{figure}[H]
    \centering
    \begin{subfigure}[b]{0.45\textwidth}
         \centering
         \includegraphics[width=\textwidth]{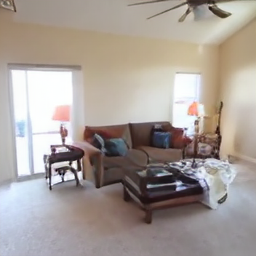}
         \caption{Initial image}
     \end{subfigure}
     \hfill
     \begin{subfigure}[b]{0.45\textwidth}
         \centering
         \includegraphics[width=\textwidth]{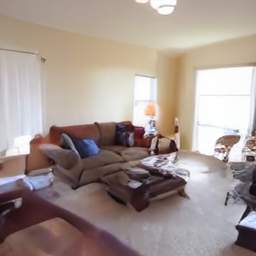}
         \caption{Image after returning close to the initial camera position.}
     \end{subfigure}
    \caption{The initial frame and the final frame from a generated sequence with the \textit{spin} motion.
    Notice the final frame has returned to a location similar to the initial frame, but the bottom left region on the floor has changed appearance.}
    \label{fig:no_cycle}
\end{figure}

Our method and the baselines are limited by the use of sequential generation with a fixed budget for conditioning images.
Regions that become occluded and subsequently disoccluded in a sequence are very likely to change appearance.
For example, conditioning on one image prevents information about previously disoccluded regions from informing the generation of those same regions beyond one frame.
Qualitative examples of this phenomenon can be seen on our project page, \webpage, under the \textbf{RealEstate10K Qualitative Results - Out-of-Distribution Trajectories} section, with the \textbf{Spin} motion.
The described phenomenon can be observed at the edges of the images with \textbf{Spin} motion,
where those regions of the scene often move beyond the image boundaries before returning in the future.
A qualitative example of this is shown in Figure \ref{fig:no_cycle}.

Conditioning on an arbitrary number of frames could theoretically solve this problem.
However, the practicality of this solution is limited by the ability to design models that can process an arbitrary number of inputs, and the model's ability to generalize to out-of-distribution camera poses (\textit{e.g.}, far away cameras in large scenes).
Leveraging many images for generation is a potentially significant direction for future work.

\end{document}